\documentclass[11pt]{article}

\usepackage[preprint]{acl}

\usepackage{times}
\usepackage{latexsym}

\usepackage{tabularx}
\usepackage{booktabs}

\usepackage[T1]{fontenc}

\usepackage[utf8]{inputenc}

\usepackage{microtype}

\usepackage{inconsolata}

\usepackage{graphicx}

\usepackage[utf8]{inputenc}
\usepackage[T1]{fontenc}
\usepackage{hyperref}
\usepackage{url}
\usepackage{booktabs}
\usepackage{amsfonts}
\usepackage{nicefrac}
\usepackage{microtype}
\usepackage{xcolor}
\usepackage{wrapfig}
\usepackage{mdframed}

\usepackage{amsthm}
\newtheorem{theorem}{Theorem}[section]

\usepackage[textsize=tiny]{todonotes}
\usepackage[most]{tcolorbox}

\usepackage{amsmath,amsfonts,bm}

\def\eqref#1{equation~\ref{#1}}

\def\1{\bm{1}}

\def\vtheta{{\bm{\theta}}}

\def\ve{{\bm{e}}}

\def\vr{{\bm{r}}}

\def\vx{{\bm{x}}}
\def\vy{{\bm{y}}}

\DeclareMathAlphabet{\mathsfit}{\encodingdefault}{\sfdefault}{m}{sl}
\SetMathAlphabet{\mathsfit}{bold}{\encodingdefault}{\sfdefault}{bx}{n}

\def\gD{{\mathcal{D}}}
\def\gE{{\mathcal{E}}}

\def\gL{{\mathcal{L}}}
\def\gM{{\mathcal{M}}}

\def\gO{{\mathcal{O}}}

\def\gR{{\mathcal{R}}}

\def\sB{{\mathbb{B}}}

\def\sD{{\mathbb{D}}}

\newcommand{\E}{\mathbb{E}}

\DeclareMathOperator*{\argmin}{arg\,min}

\title{Explanation-Guided Distillation for Human-Aligned LLM-Judges}

\author{%
  \textbf{Pouya M. Ghari\textsuperscript{1}},
  \begingroup
    \setcounter{footnote}{1}%
    \textbf{Heinrich Peters\textsuperscript{1}\thanks{\;Corresponding author: \textit{heinrich\_peters2@apple.com}}}%
  \endgroup,  \textbf{Jialu Wang\textsuperscript{1}},
  \textbf{Navid Hashemi\textsuperscript{1}},
  \textbf{Alireza Hashemi\textsuperscript{1}}, \\
  \textbf{Asad A. Butt\textsuperscript{1}},
  \textbf{Joseph Hoover\textsuperscript{1}},
  \textbf{Morteza Dehghani\textsuperscript{1}},
  \textbf{James R. Rae\textsuperscript{1}} \\
\\
  \textsuperscript{1}Apple\\
}

\begin{document}

\maketitle

\begin{abstract}

Large language models (LLMs) are increasingly used as automated evaluators designed to emulate human judgment, yet reliably matching human ratings remains a challenge. Prompted LLM judges often lack task-specific supervision, while specialized reward models are prone to overfitting. We propose \emph{Exp-Eval}, a framework that combines the strengths of both approaches by distilling structured, criterion-level explanations from a strong LLM judge into lightweight models trained on task-specific supervised data. The system decomposes evaluation into two components: a rating module that predicts human scores and an explainer module that generates structured evidence to support those predictions.
To promote the usefulness of the explanations and encourage exploration of the explanation space, Exp-Eval introduces an RLHF-based training procedure for the explainer module. The explainer is first fine-tuned on teacher-generated explanations and then refined via RLHF, using a reward signal provided by the rating module, which is constructed based on observed ratings in the supervised data.
To improve robustness under distribution shift, we introduce an online active ensembling strategy that dynamically combines the rating module's scores with the teacher's ratings under a label query-budget, with provable sublinear regret guarantees.
Experiments across diverse domains show that Exp-Eval improves correlation with human judgments over strong LLM-as-a-judge baselines while maintaining efficiency. These results demonstrate that Exp-Eval offers a flexible, efficient, and human-aligned solution for scalable system evaluation.

\end{abstract}

\section{Introduction}
AI systems increasingly rely on automated evaluation to scale development. Because human evaluation is too slow and expensive for rapid iteration, large foundation models are routinely used as evaluators, a practice termed \emph{LLM-as-a-judge} \cite{Clark2021,Zheng2023,Liu2023geval}. However, reliably matching human ratings remains challenging as human preferences are noisy, heterogeneous, and sensitive to subtle aspects of output quality.

Most prior work follows one of two directions. First, prompting frontier models \cite{Dubois2023,Kim2024,Saha2024,Wang2024self,Trivedi2024,Yuan2024,Zhang2024reference,Li2025scoring,Dsouza2025}, which have broad world-knowledge but do not incorporate task-specific training. Second, training reward models on preference data \cite{Chen2021faithful,Hou2024chatglm}, which can adapt to specific domains, but learning a single scalar function from limited data can be brittle and prone to distribution shift.

To overcome these complementary limitations, this paper introduces \textbf{Exp-Eval}, an evaluation framework that combines the task-specific calibration of lightweight models with the broad prior knowledge of frontier LLM judges. Our central contribution is a structured distillation approach, where the outputs of a large teacher model are distilled into two cooperating lightweight components: an \emph{explainer module} that generates structured evidence, and a \emph{rating module} that predicts human ratings conditioned on these explanations.

While standard supervised fine-tuning (SFT) can teach a model to generate explanations, it often fails to explore the diverse space of possible rationales to find those most useful for predicting the actual rating. To address this problem, we propose a training method for the explainer module using Reinforcement Learning from Human Feedback (RLHF) \cite{Ouyang2022}, enabling it to collaborate with the rating module and learn to generate better explanations.
Importantly, we use the rating module itself to compute the reward. By rewarding explanations that maximize the likelihood of the observed ratings, the explainer learns to generate evidence tailored specifically to support the rating module's performance.

\begin{figure*}[t]
    \centering
    \includegraphics[width=0.9\textwidth]{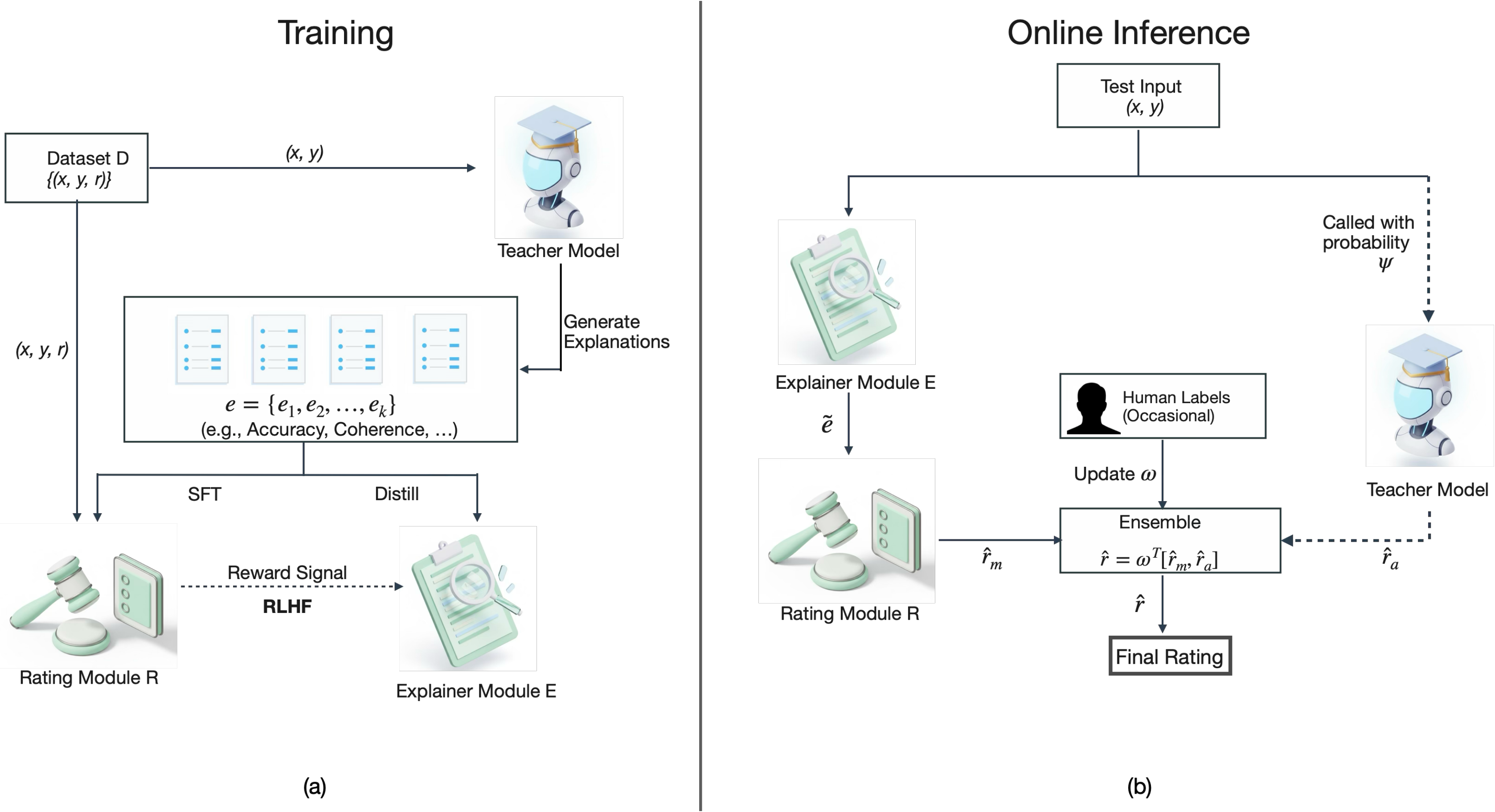}
    \caption{\textbf{Overview of Exp-Eval.}
    (a) \textbf{Training:} A teacher model (e.g., a frontier LLM) generates structured explanations $\ve = \{e_1, e_2, \ldots, e_k\}$ for an input–output pair $(\vx, \vy)$ from the dataset $\gD := {(\vx, \vy, r)}$, where $r$ denotes the rating corresponding to $(\vx, \vy)$. The Rating Module $\mathcal{R}$ is trained via SFT to predict ratings $r$ given $(\vx, \vy, \ve)$. The Explainer Module $\mathcal{E}$ is first distilled from the teacher explanations and then refined via RLHF, using $\mathcal{R}$ as the reward model (Eqs.~\ref{eq:4}–\ref{eq:6}).
    (b) \textbf{Online Inference:} The Explainer Module $\mathcal{E}$ generates explanations $\tilde{\ve}$ for test inputs, which are passed to the Rating Module $\mathcal{R}$ to produce the rating $\hat{r}_m$. With probability $\psi$ (Eq.~\ref{eq:9}), the teacher model is queried to obtain the rating $\hat{r}_a$. The ratings are linearly combined using learned ensemble weights (Eq.~\ref{eq:ens_rating}). Occasional human labels are used to update the ensemble weights online (Eqs.~\ref{eq:weight_update}–\ref{eq:12}), enabling adaptive performance without retraining.}
    \label{fig:overview}
\end{figure*}

Finally, because lightweight evaluators and frontier judges have complementary strengths (task-specific calibration vs. robustness), we propose an online active ensembling strategy for inference. This mechanism adaptively combines their ratings under a human label query budget, yielding provable sublinear regret guarantees to approach the performance of the best fixed ensemble (Figure~\ref{fig:overview}).

We evaluate Exp-Eval across diverse domains, including text summarization, AI assistant safety, personalized recommendation, and AI-generated image evaluation. Across all tasks, Exp-Eval tends to achieve stronger correlations with human judgments than strong LLM-as-a-judge baselines while maintaining inference efficiency. The main contributions of this paper are:

\begin{itemize}
    \item \textbf{The Exp-Eval Framework:} An evaluation method that distills structured, criterion-level explanations from a frontier teacher into lightweight, collaborating explainer and rating modules.
    \item \textbf{Rating-Guided RLHF:} A training procedure that optimizes the explainer module to generate rationales that explicitly improve the rating module's predictive performance.
    \item \textbf{Online Active Ensembling:} A robust, budget-aware inference strategy with theoretical regret guarantees that dynamically combines specialized lightweight models with generalist teachers.
\end{itemize}

\section{Related Work}

\textbf{LLM-as-a-judge and automatic evaluation}
LLMs have been widely adopted as automated evaluators to replace or supplement human judgment \cite{Clark2021,Zheng2023,Liu2023geval}. In this paradigm, an LLM is prompted to assign a score, select a preferred output, or generate a critique of a system response \cite{Dubois2023,Kim2024}. A substantial body of work has focused on improving the reliability of such evaluations through structured prompting, including fine-grained criteria \cite{Saha2024,Wang2024self,Trivedi2024}, explicit rubrics \cite{Yuan2024,Li2025scoring,Dsouza2025,Xu2026alternating,Kawabata2026}, and the use of reference answers or self-improvement loops \cite{Zhang2024reference,Saha2025}. In contrast, our approach uses a strong LLM as a \emph{teacher} to generate structured explanations, which are distilled into a task-adapted, lightweight evaluator in order to use the efficiency advantages of lightweight models for agentic applications \cite{Belcak2025small}.

\textbf{Reward modeling and preference learning}
Another line of work trains reward models from human feedback to produce scores for evaluating system outputs \cite{Ouyang2022,Chen2021faithful,Hou2024chatglm}. These models are typically optimized on pairwise preferences or rankings and have been extended with critics, reasoning traces, or multi-stage architectures to improve expressivity \cite{Rame2023,Akner2024,Yu2025,Xie2025teaching,Chen2025rmr1}. Most reward models are trained to capture relative preference rather than absolute ratings, which are common in practical feedback settings (e.g., Likert or star ratings). Exp-Eval differs in that it explicitly targets rating prediction and uses explanation-based intermediate supervision to make this learning problem easier and more stable.

A supplementary discussion of related work is provided in Appendix \ref{supp:related_works}.

\section{Proposed Evaluation with Explanations}
The problem of designing an evaluator model can be formulated as learning a function $f(\cdot,\cdot)$ that, given an input $\vx$ and an output $\vy$, evaluates the performance of the system by assigning a rating $r = f(\vx,\vy)$ corresponding to $\vx$ and $\vy$. When employing LLMs as evaluators, this problem reduces to training a policy $\pi_\vtheta$, parameterized by $\vtheta$, for the LLM model $\gM$ to sample a rating $r \sim \pi_\vtheta(\vx, \vy)$ given an input prompt $\vx$ and an output response $\vy$. Let there be a dataset of ratings $\sD := \{(\vx_i, \vy_i, r_i)\}$, where $r_i$ is sampled from an unknown true rating distribution $p(r_i|\vx_i, \vy_i)$. This setting arises, for example, when human ratings from users or annotators are available for a dataset of prompt–response pairs. Due to uncertainty in human ratings as well as diversity among humans, the label is sampled from a stochastic distribution rather than being deterministic. Given the rating dataset $\sD$, the goal is to fine-tune the LLM to obtain parameters $\vtheta$ such that the LLM policy $\pi_\vtheta$ samples ratings with minimum discrepancy from the given rating labels $\{r_i\}$.

\subsection{Learning from Explanations}
Let $\hat{r}_i$ be the rating sampled by $\pi_\vtheta$ corresponding to $\vx_i$ and $\vy_i$.
It is desirable to learn a policy $\pi_{\vtheta}(\hat{r}_i \mid \vx_i, \vy_i)$ that captures as much information as possible about $p(r_i \mid \vx_i, \vy_i)$. Introducing a latent variable $\ve_i$ may increase the information about $p(r_i \mid \vx_i, \vy_i)$ if $\ve_i$ contains non-redundant information that is not already captured by $\pi_{\vtheta}(\hat{r}_i \mid \vx_i, \vy_i)$.
Therefore, it can be concluded that sampling the estimated rating for $(\vx_i, \vy_i)$ conditioned on a latent variable $\ve_i$, i.e., $\hat{r}_i \sim \pi_\vtheta(r\mid \vx_i, \vy_i, \ve_i)$, can improve the performance of the evaluator provided that the latent variable $\ve_i$ contains information about $p(r_i\mid \vx_i,\vy_i)$ that is not present in either the rating dataset or the prior knowledge of the LLM model $\gM$. One approach to obtaining a latent variable $\ve_i$ that contains novel and non-redundant information is to acquire such information from external sources, referred to as \emph{teacher models}. In this context, teacher models can provide $\ve_i$ in the form of explanations about the prompt–response pair $(\vx_i,\vy_i)$ to quantify the performance of the system. The teacher model can be an LLM with higher or different expertise than the LLM $\gM$ with the policy $\pi_\vtheta$. To improve evaluation efficiency, this paper considers the case in which the LLM $\gM$ is a lightweight model.
To provide effective explanations, information should be extracted from the teacher models that is relevant to the evaluation task and can be provided reliably by the teacher. To this end, we first use the teacher to structure the evaluation by decomposing the task into relevant sub-dimensions or evaluation criteria. This can be achieved by prompting the teacher model to extract these criteria. Then, for each prompt-response pair $(\vx_i, \vy_i)$ in the dataset, the teacher evaluates the pair based on each criterion, and all these evaluations are collected into the explanation $\ve_i$. Such a multi-dimensional evaluation facilitates a better understanding of the relationship between the observed rating $r_i$ and the prompt–response pair $(\vx_i, \vy_i)$.

After generating explanations for the data samples, the next step is to train the LLM policy $\pi_\vtheta$ to sample ratings $\hat{r}_i \sim \pi_\vtheta(r|\vx_i, \vy_i, \ve_i)$, which is then used as the rating module during inference. To train the rating module $\gR$, the LLM policy $\pi_\vtheta$ is fine-tuned through SFT on the rating labels using the dataset $\{(\vx_i, \vy_i, \ve_i)\}$. In this context, the observed ratings $\{r_i\}$ in the dataset are converted into textual annotations and the LLM is fine-tuned to auto-regressively generate the rating.

\subsection{Generating Explanations} \label{sec:gen-exp}
As discussed in the previous section, the role of explanations is to assist the rating module $\gR$ in generating ratings that are better aligned with human judgments. Consequently, learning to generate explanations that are useful to the rating module $\gR$ is a crucial problem. To address this, we train an explainer module $\gE$ using the LLM $\gM$ to support the rating module by producing helpful explanations. Specifically, the explainer module first learns to generate explanations from the teacher model via SFT. It then further improves explanation through RLHF, using reward signals from the rating module that reflect the helpfulness of the generated explanations. In what follows, we discuss training the explainer module $\gE$ in detail.

Let $\pi_{\boldsymbol{\rho}}$ denote the LLM policy parameterized by $\boldsymbol{\rho}$, trained from the base LLM $\gM$ to serve as the explainer module $\gE$. The explanations generated by $\gE$ are subsequently provided to the rating module $\gR$ to produce the final rating. To enable the explainer module to learn the additional information conveyed by the teacher, we fine-tune $\pi_{\boldsymbol{\rho}}$ via SFT on the explanations ${\ve_i}$ generated by the teacher for the input–output pairs ${(\vx_i, \vy_i)}$.
Notably, the primary goal of the explanations is to assist the rating module $\gR$ in generating ratings that most closely agree with the observed ratings in the dataset. A natural next step is to use the ratings $\{r_i\}$ in the dataset to further improve the explanations, such that they provide more effective assistance to the rating module $\gR$. To achieve this, the explainer module $\gE$ can be fine-tuned via RLHF to produce improved explanations. Policy gradient RLHF methods such as PPO \cite{Schulman2017proximal} and GRPO \cite{Shao2024deepseekmath} can be employed to fine-tune the explainer module, where the policy is updated using gradient descent. At each optimization step $t$, a batch of prompt–response pairs $\sB_t$ is sampled from the dataset $\sD$. Let $\boldsymbol{\rho}_t$ represent the policy parameters at optimization step $t$. Furthermore, let $\tilde{\ve}_{i,t}$ denote the explanation generated by $\pi_{\boldsymbol{\rho}_t}$ at step $t$ for $(\vx_i, \vy_i)$, where $(\vx_i, \vy_i) \in \sB_t$. The reward of generating the explanation $\tilde{\ve}_{i,t}$ for $(\vx_i, \vy_i)$ is obtained as:
\begin{align}
    R(\tilde{e}_{i,t}|\vx_i, \vy_i) = \log \pi_\vtheta(r_i|\vx_i,\vy_i,\tilde{\ve}_{i,t}). \label{eq:4}
\end{align}
To bound the reward values, $\sigma(\log \pi_\vtheta(r_i|\vx_i,\vy_i,\tilde{\ve}_{i,t}))$ can instead be used as the reward where $\sigma(\cdot)$ denotes an activation function.
Given the loss function $\gL(\cdot)$ specified by the RLHF method and the reward $R(\tilde{e}_{i,t}|\vx_i, \vy_i)$, $\forall (\vx_i, \vy_i) \in \sB_t$, the explainer policy $\boldsymbol{\rho}_t$ is updated as:
\begin{align}
    \boldsymbol{\rho}_{t+1} = \boldsymbol{\rho}_t - \frac{\eta}{|\sB_t|}\sum_{(\vx_i,\vy_i)\in \sB_t}{\nabla \gL(R(\tilde{e}_{i,t}|\vx_i, \vy_i),\boldsymbol{\rho}_t)}. \label{eq:6}
\end{align}
According to \eqref{eq:4} and \eqref{eq:6}, it can be concluded that the rating policy $\pi_\vtheta$ guides the explainer policy $\pi_{\boldsymbol{\rho}}$ via gradient descent to generate explanations that can reduce the uncertainty of the rating policy and help the rating policy to generate observed ratings in the dataset with higher likelihood.

It is noteworthy that, since the rating policy $\pi_\vtheta$ samples ratings conditioned on explanations, the explainer policy $\pi_{\boldsymbol{\rho}}$ affects the distribution from which ratings are sampled by $\pi_\vtheta$. Appendix \ref{sec:analysis-reward} relates the expected reward in \eqref{eq:4} to the Kullback--Leibler (KL) divergence between the true distribution $p$ and the rating policy $\pi_\vtheta$. Theorem \ref{th:2} shows that a smaller KL divergence corresponds to a larger upper bound on the expected reward. Since $p$ is unknown and the KL divergence cannot be minimized directly, the reward in \eqref{eq:4} serves as a practical likelihood-based proxy motivated by this bound.

\subsection{Online Ensemble Learning} \label{sec:end}
Adaptation to dynamic distribution shifts at inference time is important. Owing to their strong and general prior knowledge, teacher models can provide more robustness to distribution shifts when encountering novel prompt–response pairs while the explainer and rating policies $\pi_{\boldsymbol{\rho}}$ and $\pi_\vtheta$ can use task-specific knowledge acquired during training at inference time. The teacher and the trained policies $\pi_{\boldsymbol{\rho}}$ and $\pi_\vtheta$ therefore offer complementary advantages at inference time, implying that it might be beneficial to ensemble them `on the fly'. This paper presents an online ensemble learning approach, detailed in Appendix \ref{sec:ens-sm}, for constructing rating predictions by adaptively and linearly combining the teacher model’s ratings with those of the rating module. The ensemble algorithm learns the combination weights through occasional feedback from human annotators. We prove that the algorithm achieves sublinear regret with respect to the best ensemble model, meaning that, in the long run, its performance achieves that of the best ensemble in hindsight. Theorem \ref{th:1} in Appendix \ref{sec:ens-sm} provides a formal analysis of the algorithm’s regret.
\section{Experiments} \label{sec:exp}
Here, we evaluate Exp-Eval on diverse datasets, benchmarking multiple model variants against prompted LLM judges and reward-model baselines to assess how closely its judgments correlate with human ratings.

\subsubsection*{Datasets}
To assess the alignment of Exp-Eval ratings with human ratings, we conduct experiments on five distinct datasets that encompass diverse evaluation challenges, ranging from objective content accuracy to subjective user preferences.\\
\textbf{Summarize from Feedback:} This dataset contains original texts paired with their corresponding summaries. The texts are either Reddit posts or news articles. The dataset contains human judgments of summary accuracy \cite{Stienon2020}.\\
\textbf{HH-RLHF:} This dataset contains conversations between humans and an AI assistant. The evaluation task measures the success of a human adversary in eliciting harmful instructions from the assistant. Human annotators rate the adversary’s success \cite{Ganguli2022}.\\
\textbf{Goodreads:} This dataset includes book reviews and ratings provided by Goodreads users \cite{Wan2018,Wan2019}. The task is to predict user satisfaction with a book in order to evaluate the performance of personalized recommender systems.\\
\textbf{GenImageNet:} This dataset contains images generated by image generation models given a prompt. The task is to assess the quality of the images. The dataset contains human evaluations of each image’s overall quality \cite{Hartmann2025}.\\
\textbf{RewardBench}: To test Exp-Eval’s generalizability beyond its training tasks, we evaluate it on the prominent evaluation dataset, RewardBench 2 \cite{Lambert2025, malik2026rewardbench}.\\
More details about the datasets and tasks are provided in Appendix \ref{supp:exp}.

\begin{table}[t!]
\centering
\caption{Performance on Text Summarization in terms of Spearman correlation ($\rho$) and Kendall's $\tau$. The backbone model used for Exp-Eval, and the
reward models is Gemma-2-9B.}
\label{tab:text_summarization}
\small
\setlength{\tabcolsep}{16pt}
\begin{tabular}{lcc}
\toprule
Method & $\rho$ & $\tau$ \\
\midrule
GPT-OSS-120b & 0.4388 & 0.3737 \\
Qwen-3-VL-32B & 0.3355 & 0.2893 \\
Claude-Sonnet-4.5 & 0.4636 & 0.3987 \\
GenRM-CoT & 0.3578 & 0.2996 \\
GenRM & 0.2461 & 0.2323 \\
\midrule
Exp-Eval-SFT & \underline{0.4641} & \underline{0.4256} \\
Exp-Eval-RLHF & \textbf{0.4804} & \textbf{0.4336} \\
\bottomrule
\end{tabular}
\end{table}

\begin{table*}[t!]
\centering
\caption{Comparison of Exp-Eval-RLHF against baselines on the text summarization task. Differences in correlation ($\Delta\rho$, $\Delta\tau$) are reported with 95\% confidence intervals derived via bootstrapping. Win rates indicate the proportion of bootstrap samples where Exp-Eval-RLHF achieves a higher correlation than the base.}
\label{tab:6}
\small
\setlength{\tabcolsep}{14pt}
\begin{tabular}{l||cc|cc}
\toprule
 & \multicolumn{2}{c}{Spearman} & \multicolumn{2}{c}{Kendall} \\
Method & $\Delta \rho$ & win rate (\%) & $\Delta \tau$ & win rate (\%) \\
\hline
GPT-OSS-120b & $[+0.0073, +0.0590]$ & 99.52 & $[+0.0297, +0.0766]$ & 100.00 \\
Qwen-3-VL-32B & $[+0.1110, +0.1680]$ & 100.00 & $[+0.1142, +0.1655]$  & 100.00 \\
Claude-Sonnet-4.5 & $[-0.0081, +0.0442]$ & 90.87 & $[+0.0130, +0.0605]$ & 99.91 \\
\hline
GenRM-CoT & $[+0.0979, +0.1481]$ & 100.00 & $[+0.1145, +0.1591]$ & 100.00  \\
GenRM & $[+0.2056, +0.2783]$ & 100.00  & $[+0.1742, +0.2416]$  & 100.00  \\
\bottomrule
\end{tabular}
\end{table*}

\begin{figure}[t]
    \centering
    \includegraphics[width=1\columnwidth]{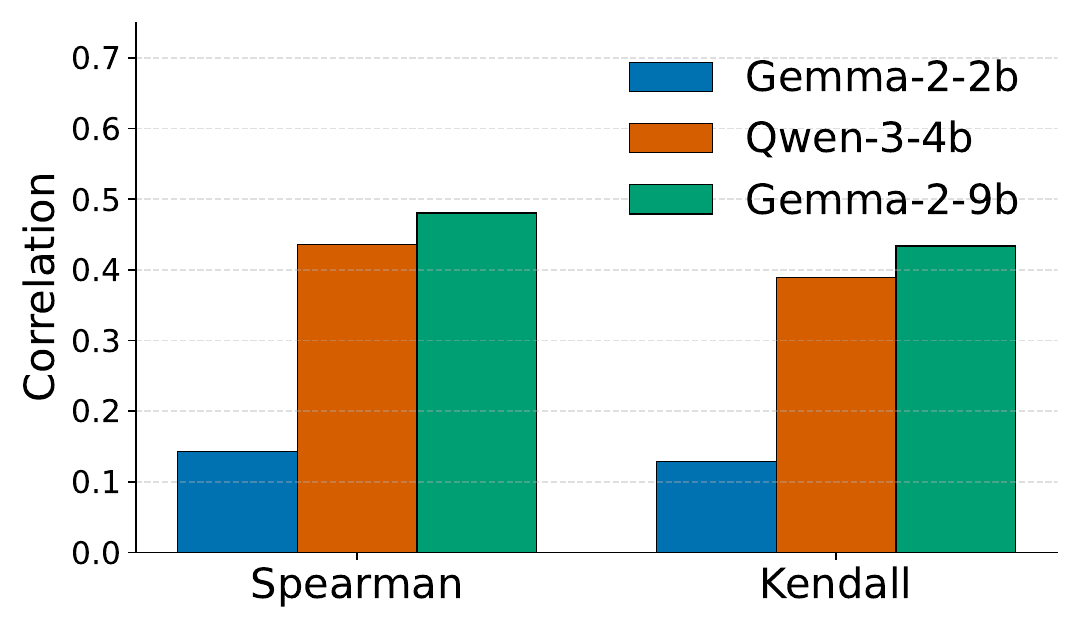}
    \caption{Comparison of Exp-Eval using three different model variants on the text summarization task.}
    \label{fig:ffs_models}
\end{figure}

\subsubsection*{Methods \& Metrics}  We compare the performance of the proposed Exp-Eval framework against both LLM-as-a-judge and reward-model baselines. Specifically, we use GPT-OSS-120B \cite{openai2025gptoss120b} and Qwen-3-VL-32B \cite{qwen3technicalreport} as open-source LLM judges to assign scores for each evaluation task. Furthermore, we employ Claude-Sonnet-4.5 as a closed-source LLM judge.
In addition, Exp-Eval is compared against the reward-model baselines GenRM and GenRM-CoT \cite{Zhang2025}. GenRM generates ratings directly from an input–output pair without reasoning or explanation, whereas GenRM-CoT produces chain-of-thought reasoning before concluding with a rating.

For text generation applications, the rating and explainer modules of Exp-Eval are trained on the \textit{Summarize from Feedback}, \textit{HH-RLHF}, and \textit{Goodreads} datasets.
We implement Exp-Eval using three lightweight LLMs: Gemma-2-2B \cite{gemma2024, gemma2}, Qwen-3-4B \cite{qwen3technicalreport}, and Gemma-2-9B \cite{gemma2024, gemma2}. We use GPT-OSS-120B as the teacher model. For each backbone model, the rating module is implemented as a LoRA adapter \cite{Hu2022lora} trained via SFT on the three datasets combined. The explainer module is implemented as a separate LoRA adapter using GRPO. To further align the rating module with the explainer, we train an additional LoRA adapter for the rating module via SFT using explanations generated by the explainer.
For the image generation evaluation task, we implement Exp-Eval using Gemma-3-4B \cite{gemma3} as the student model and Qwen-3-VL-32B as the teacher model. The rating and explainer modules are trained on the \textit{GenImageNet} dataset. The remaining implementation is similar to that used for text generation applications. Results on the RewardBench2 dataset are presented in Appendix \ref{supp:results}.
Further implementation details of Exp-Eval and the baseline methods are provided in the Appendix \ref{supp:exp}.
The performance of Exp-Eval and the baseline methods is evaluated using Spearman and Kendall correlations between predicted ratings and ground-truth human ratings on the test datasets.

\begin{table}[t]
\centering
\caption{Performance on AI Assistant Safety in terms of Spearman correlation ($\rho$) and Kendall's $\tau$. The backbone model used for Exp-Eval, and the reward models is Gemma-2-9B.}
\label{tab:ai_safety}
\small
\setlength{\tabcolsep}{16pt}
\begin{tabular}{lcc}
\toprule
Method & $\rho$ & $\tau$ \\
\midrule
GPT-OSS-120b & 0.6418 & 0.5570 \\
Qwen-3-VL-32B & 0.3130 & 0.2736 \\
Claude-Sonnet-4.5 & 0.6383 & 0.5224 \\
GenRM-CoT & 0.6406 & 0.5107 \\
GenRM & \underline{0.6722} & \underline{0.5919} \\
\midrule
Exp-Eval-SFT & 0.6645 & 0.5614 \\
Exp-Eval-RLHF & \textbf{0.7085} & \textbf{0.6069} \\
\bottomrule
\end{tabular}
\end{table}

\subsection{Text Summarization}
First, we present the performance of Exp-Eval compared with baseline methods on the evaluation of text summarization using samples from the \textit{Summarize from Feedback} dataset. Methods that require training are trained on Reddit post summarization, while evaluation is conducted on news article summarization to assess performance on out-of-distribution summarization tasks.

The task is to rate the accuracy of summaries. Exp-Eval is trained using explanations generated by GPT-OSS-120B, where the model evaluates summaries with respect to Accuracy, Coherence, and Coverage by providing short explanations for each criterion, followed by an assigned score. Exp-Eval-RLHF uses an explainer module trained via GRPO, while Exp-Eval-SFT employs an explainer module trained via SFT.

Results in Table \ref{tab:text_summarization} show that Exp-Eval-RLHF achieves higher correlations compared to Exp-Eval-SFT, GenRM, and GenRM-CoT. A comparison between Exp-Eval-RLHF and Exp-Eval-SFT demonstrates the effectiveness of the proposed RLHF training method. Furthermore, the results indicate that Exp-Eval-RLHF effectively uses explanations for rating generation, compared to GenRM, which directly generates ratings, and GenRM-CoT, which generates ratings following chain-of-thought reasoning. Results in Table \ref{tab:text_summarization} show that, while Exp-Eval-RLHF is more efficient than LLM-as-a-judge baselines due to its lightweight modules, it achieves higher correlations than GPT-OSS-120B, Qwen-3-VL-32B, and Claude-Sonnet-4.5. To further explore the relative performance of the baselines compared to Exp-Eval-RLHF, we conduct statistical significance tests via bootstrapping, in which the dataset is resampled 10,000 times as described in Appendix \ref{supp:boot}. For each resample, we compute the difference between the correlation of Exp-Eval-RLHF and that of each baseline. Table \ref{tab:6} presents the 95\% confidence intervals for differences in correlations, along with Exp-Eval-RLHF’s win rates relative to the baselines. These results show that Exp-Eval-RLHF tends to outperform the baselines. Figure \ref{fig:ffs_models} illustrates the performance of Exp-Eval-RLHF using lightweight model variants of different sizes on the text summarization task. As shown, increasing model size improves Exp-Eval’s performance.

\begin{table*}[t]
\centering
\caption{Comparison of Exp-Eval-RLHF against baselines on the AI Assistant Safety task. Differences in correlation ($\Delta\rho$, $\Delta\tau$) are reported with 95\% confidence intervals derived via bootstrapping. Win rates indicate the proportion of bootstrap samples where Exp-Eval-RLHF achieves a higher correlation than the base.}
\label{tab:10}
\small
\setlength{\tabcolsep}{14pt}
\begin{tabular}{l||cc|cc}
\toprule
 & \multicolumn{2}{c}{Spearman} & \multicolumn{2}{c}{Kendall} \\
Method & $\Delta \rho$ & win rate (\%) & $\Delta \tau$ & win rate (\%) \\
\hline
GPT-OSS-120b & $[+0.0544, +0.0790]$ & 100.00 & $[+0.0385, +0.0611]$ & 100.00 \\
Qwen-3-VL-32B & $[+0.3709, +0.4166]$ & 100.00 & $[+0.3116, +0.3520]$  & 100.00 \\
Claude-Sonnet-4.5 & $[+0.0510, +0.0801]$ & 100.00 & $[+0.0679, +0.0942]$ & 100.00 \\
\hline
GenRM-CoT & $[+0.0568, +0.0795]$ & 100.00 & $[+0.0865, +0.1075]$ & 100.00  \\
GenRM & $[+0.0260, +0.0432]$ & 100.00  & $[+0.0060, +0.0219]$  & 99.96  \\
\bottomrule
\end{tabular}
\end{table*}

\subsection{AI Assistant Safety}
Next, we evaluate Exp-Eval on the \textit{HH-RLHF} dataset to assess its performance in rating the success of human adversaries attempting to elicit harmful instructions from an AI assistant. We split the data into 70\% training, 10\% validation, and 20\% test sets, and use the training split whenever training is required.

Table \ref{tab:ai_safety} reports Spearman and Kendall correlations for the ratings produced by each method. All experimental settings follow those used in the previous task.
The results show that methods relying on the training data tend to achieve higher correlations than LLM-as-a-judge baselines on this task. Furthermore, while Exp-Eval-SFT and GenRM-CoT do not achieve higher correlations than GenRM, Exp-Eval-RLHF achieves significantly higher correlations than all baselines, indicating the effectiveness of the proposed RLHF method in exploring the space of explanations. Table \ref{tab:10} presents the 95\% confidence intervals for the correlation differences between Exp-Eval-RLHF and the other baselines, along with Exp-Eval-RLHF’s win rates. These results confirm that Exp-Eval-RLHF outperforms the baselines on the AI Assistant Safety task.

\begin{table}[t]
\centering
\caption{Performance on Personalized Recommendation in terms of Spearman correlation ($\rho$) and Kendall's $\tau$. The backbone model used for Exp-Eval, and the reward models is Gemma-2-9B.}
\label{tab:personalization}
\small
\setlength{\tabcolsep}{16pt}
\begin{tabular}{lcc}
\toprule
Method & $\rho$ & $\tau$ \\
\midrule
GPT-OSS-120b & 0.1993 & 0.1537 \\
Qwen-3-VL-32B & 0.2385 & 0.1954 \\
Claude-Sonnet-4.5 & 0.2050 & 0.1651 \\
GenRM-CoT & 0.1391 & 0.1082 \\
GenRM & \underline{0.2855} & \underline{0.2553} \\
\midrule
Exp-Eval-SFT & 0.2512 & 0.2165 \\
Exp-Eval-RLHF & \textbf{0.3462} & \textbf{0.2882} \\
\bottomrule
\end{tabular}
\end{table}

\begin{table*}[t]
\centering
\caption{Comparison of Exp-Eval-RLHF against baselines on the personalized recommendation task. Differences in correlation ($\Delta\rho$, $\Delta\tau$) are reported with 95\% confidence intervals derived via bootstrapping. Win rates indicate the proportion of bootstrap samples where Exp-Eval-RLHF achieves a higher correlation than the base.}
\label{tab:11}
\small
\setlength{\tabcolsep}{14pt}
\begin{tabular}{l||cc|cc}
\toprule
 & \multicolumn{2}{c}{Spearman} & \multicolumn{2}{c}{Kendall} \\
Method & $\Delta \rho$ & win rate (\%) & $\Delta \tau$ & win rate (\%) \\
\hline
GPT-OSS-120b & $[+0.1025, +0.1911]$ & 100.00 & $[+0.0976, +0.1691]$ & 100.00 \\
Qwen-3-VL-32B & $[+0.0709, +0.1443]$ & 100.00 & $[+0.0606, +0.1222]$  & 100.00 \\
Claude-Sonnet-4.5 & $[+0.0984, +0.1773]$ & 100.00 & $[+0.0862, +0.1513]$ & 100.00 \\
\hline
GenRM-CoT & $[+0.1597, +0.2479]$ & 100.00 & $[+0.1411, +0.2125]$ & 100.00  \\
GenRM & $[+0.0320, +0.0913]$ & 100.00  & $[+0.0064, +0.0582]$  & 99.25  \\
\bottomrule
\end{tabular}
\end{table*}

\subsection{Personalized Recommendation}
In this section, we evaluate Exp-Eval alongside baseline methods on a personalized recommendation task using the \textit{Goodreads} dataset. Given a user and a book, each paired with their respective review histories, the objective is to predict the user’s specific rating for that book. To perform this prediction, each method is provided with a summary of the user’s reading preferences extracted from the user’s history, along with a summary of the book’s characteristics derived from its previous reviews and ratings. GPT-OSS-120B is used to generate these summaries from the user and book histories. We extract 50,000 user–book pairs with review histories and use 90\% of them for training and the remaining 10\% for validation. Note that the training and validation samples are not used for summary generation. For testing, both users and books are partitioned into seen and unseen sets. A user or book is considered seen if it appears at least once in the training set; otherwise, it is considered unseen.

Tables \ref{tab:personalization} and \ref{tab:11} report results for pairs involving unseen users and unseen books. Exp-Eval-RLHF achieves significantly higher correlations than all baselines.

\begin{table}[t]
\centering
\caption{Performance on Image Generation Quality in terms of Spearman correlation ($\rho$) and Kendall's $\tau$.
}
\label{tab:4}
\small
\setlength{\tabcolsep}{15pt}
\begin{tabular}{l||cc}
\toprule
Method & $\rho$ & $\tau$ \\
\hline
Qwen-3-VL-32B & 0.2567 & 0.1881 \\
Claude-Sonnet-4.5 & 0.2610 & 0.1837 \\
GenRM-CoT & 0.2655 & 0.1791 \\
GenRM & 0.1987 & 0.1356 \\
\hline
Exp-Eval-SFT & \underline{0.2870} & \underline{0.1949} \\
Exp-Eval-RLHF & \textbf{0.3147} & \textbf{0.2132} \\
Exp-Eval-ensemble & \textbf{0.3358} & \textbf{0.2263} \\
\bottomrule
\end{tabular}
\end{table}

\begin{table*}[t]
\centering
\caption{Comparison of Exp-Eval-RLHF against baselines on the image generation quality assessment task. Differences in correlation ($\Delta\rho$, $\Delta\tau$) are reported with 95\% confidence intervals derived via bootstrapping. Win rates indicate the proportion of bootstrap samples where Exp-Eval-RLHF achieves a higher correlation than the base.}
\label{tab:12}
\small
\setlength{\tabcolsep}{14pt}

\begin{tabular}{l||cc|cc}
\toprule
 & \multicolumn{2}{c}{Spearman} & \multicolumn{2}{c}{Kendall} \\
Method & $\Delta \rho$ & win rate (\%) & $\Delta \tau$ & win rate (\%) \\
\hline
Qwen-3-VL-32B & $[-0.0063, +0.1223]$ & 96.29 & $[-0.0213, +0.0718]$  & 86.42 \\
Claude-Sonnet-4.5 & $[-0.0107, +0.1291]$ & 94.87 & $[-0.0153, +0.0834]$ & 90.62 \\
\hline
GenRM-CoT & $[-0.0094, +0.1076]$ & 94.83 & $[-0.0066, +0.0747]$ & 94.83  \\
GenRM & $[+0.0455, +0.1860]$ & 99.89  & $[+0.0290, +0.1261]$  & 99.82  \\
\bottomrule
\end{tabular}
\end{table*}

\begin{table*}[t]
\centering
\caption{Comparison of Exp-Eval-ensemble against baselines on the image generation quality assessment task. Differences in correlation ($\Delta\rho$, $\Delta\tau$) are reported with 95\% confidence intervals derived via bootstrapping. Win rates indicate the proportion of bootstrap samples where Exp-Eval-ensemble achieves a higher correlation than the base.}
\label{tab:2}
\small
\setlength{\tabcolsep}{14pt}
\vspace{-0.25cm}
\begin{tabular}{l||cc|cc}
\toprule
 & \multicolumn{2}{c}{Spearman} & \multicolumn{2}{c}{Kendall} \\
Method & $\Delta \rho$ & win rate (\%) & $\Delta \tau$ & win rate (\%) \\
\hline
Qwen-3-VL-32B & $[+0.0466, +0.1109]$ & 100.00 & $[+0.0141, +0.0614]$  & 99.95 \\
Claude-Sonnet-4.5 & $[+0.0174, +0.1361]$ & 99.43 & $[+0.0020, +0.0861]$ & 97.99 \\
\hline
GenRM-CoT & $[+0.0119, +0.1253]$ & 99.07 & $[+0.0064, +0.0851]$ & 98.86  \\
GenRM & $[+0.0679, +0.2032]$ & 99.99  & $[+0.0428, +0.1368]$  & 99.99  \\
\bottomrule
\end{tabular}
\end{table*}

\subsection{Image Generation Quality}
Here we evaluate the performance of Exp-Eval against other baselines on the quality assessment of AI-generated images. We split the images in the dataset into 80\% training, 10\% validation, and 10\% test samples. Table \ref{tab:4} presents the Spearman and Kendall correlations of Exp-Eval and other baselines on the image generation quality task. These results show that Exp-Eval-RLHF achieves the highest correlations, which indicates the effectiveness of the proposed RLHF training method. Furthermore, Table \ref{tab:12} shows the 95\% confidence intervals for the performance differences between Exp-Eval-RLHF and other baselines along with the win rates of Exp-Eval-RLHF. Although for Qwen-3-VL-32B, Claude-Sonnet-4.5, and GenRM-CoT, Exp-Eval-RLHF obtains a relatively small negative lower bound in the confidence interval, indicating on par performance, its win rates are always greater than 85\% compared to these baselines. It is noteworthy that since Exp-Eval-RLHF employs a Gemma-3-4B model, it provides considerable efficiency compared to both Qwen-3-VL-32B and Claude-Sonnet-4.5. Moreover, the lower-bound confidence interval results in Table \ref{tab:12} indicate that Qwen-3-VL-32B and Exp-Eval-RLHF may have complementary strengths, such that an ensemble of the two can provide stronger predictions. Exp-Eval-ensemble in Table \ref{tab:4} represents an online active ensemble of Qwen-3-VL-32B and Exp-Eval-RLHF, where only 10\% of samples are queried for labels from human annotators, while Qwen-3-VL-32B is queried for a rating on every sample. As observed, Exp-Eval-ensemble achieves higher correlations than Exp-Eval-RLHF. Furthermore, Table \ref{tab:2} presents the 95\% confidence intervals for the performance differences between Exp-Eval-ensemble and the other baselines, showing the effectiveness of the online active ensemble method.

\begin{table*}[!htbp]
\centering
\caption{Explanation usefulness assessment of Exp-Eval-SFT, Exp-Eval-RLHF, and perturbed Exp-Eval-RLHF explanations. Acceptance rates (\%) and 95\% confidence intervals are reported for the three evaluation tasks.}
\label{tab:8}
\small
\setlength{\tabcolsep}{24pt}
\begin{tabular}{l||c|c|c}
\toprule
Explanations & Text Summarization & AI Assistant Safety & Personalization \\
\hline
Perturbed & 68.0 $[66.8, 69.3]$ & 33.8 $[32.7, 34.8]$ & 84.6 $[83.2, 86.0]$\\
SFT & 97.7 $[97.3, 98.1]$ & 83.7 $[82.8, 84.5]$& 99.9 $[99.8, 100.0]$ \\
RLHF & 98.2 $[97.9, 98.6]$ & 84.6 $[83.8, 85.4]$ & 99.9 $[99.7, 100.0]$ \\
\bottomrule
\end{tabular}
\end{table*}

\subsection{Explanation Usefulness} \label{exp-qual}
Here, we examine the usefulness of explanations generated by Exp-Eval-RLHF. The backbone model used for explanation generation is Gemma-2-9B. For each sample in the test sets of the Summarize from Feedback, HH-RLHF, and Goodreads datasets, the explainer module generates one explanation. Qwen-3-VL-32B is employed to evaluate the usefulness of the generated explanations. Qwen-3-VL-32B is selected as the LLM judge rather than GPT-OSS-120B to ensure that the judge differs from the teacher model. We ask the LLM to either accept or reject each explanation based on its usefulness. An explanation is classified as useful only if it satisfies all predefined acceptance criteria: First, the explanation must identify evidence that is relevant to the task. Second, it must clearly connect this evidence to the task objective. Third, it must provide enough substantive information for a reader to understand the underlying evaluation. Finally, it must be specific enough to be distinguishable from a generic justification that could apply to other tasks. Explanations that are generic, superficial, non-specific to the task, or merely restate the assessment without articulating the relevant evidence are rejected. The prompt template for the usefulness assessment is provided at the end of the Appendix.

Table \ref{tab:8} presents the results of the explanation usefulness assessment for explanations generated by Exp-Eval-SFT and Exp-Eval-RLHF, as well as for perturbed explanations generated by Exp-Eval-RLHF. To assess the sensitivity of the LLM judge, we perturbed the explanation texts by replacing a subset of criterion-specific explanations with generic and nonspecific ones. Consequently, the perturbed explanations are expected to be less useful while potentially retaining some useful information from the unperturbed portions. More information about explanation perturbation protocol can be found in Appendix \ref{exp:pert}. 95\% confidence intervals for acceptance rates are computed via bootstrapping as in Appendix~\ref{supp:boot}. As shown, the acceptance rates of Exp-Eval-RLHF exceed 84\% for all three tasks, reaching 98.2\% for Text Summarization, 84.6\% for AI Assistant Safety, and 99.9\% for Personalization. These results indicate that the vast majority of generated explanations satisfy all predefined usefulness criteria according to the LLM judge. Although the primary goal of the explainer module is to assist the rating module in generating human-aligned ratings, these results demonstrate that the explainer module can also produce explanations that meet the predefined usefulness criteria. Thus, the explainer module not only helps Exp-Eval achieve higher correlations with ground-truth ratings but also produces explanations that provide useful supporting information for the evaluations. Furthermore, the usefulness of explanations generated by Exp-Eval-RLHF is roughly comparable to that of explanations generated by Exp-Eval-SFT. This suggests that, although Exp-Eval-RLHF improves correlations with ground-truth labels compared with Exp-Eval-SFT, RLHF training does not diminish their usefulness. Moreover, the decrease in usefulness observed for perturbed explanations indicates that the LLM judge can effectively detect differences in explanation usefulness.
\section{Conclusion}

In this paper, we proposed Exp-Eval, a novel framework for system evaluation that assesses output quality through explanation-guided rating. Our approach efficiently leverages both lightweight and frontier LLMs by training specialized rating and explainer modules that distill structured knowledge from a larger teacher model. Through the proposed RLHF training method, the explainer module collaborates with the rating module to generate explanations that better support the rating process. Empirical results across text summarization, AI safety, personalized recommendation, and image generation quality assessment demonstrate that Exp-Eval achieves higher correlations with ground-truth ratings than state-of-the-art LLM-as-a-judge baselines.
These findings establish a flexible and efficient solution for scalable, human-aligned system evaluation. Moving forward, this paradigm of explanation-based distillation and adaptive ensembling offers a scalable path for maintaining rigorous oversight of AI systems as they move into increasingly specialized and high-stakes domains.

\bibliography{Refereces}

@inproceedings{Yuan2024,
title={Self-Rewarding Language Models},
author={Weizhe Yuan and Richard Yuanzhe Pang and Kyunghyun Cho and Xian Li and Sainbayar Sukhbaatar and Jing Xu and Jason E Weston},
booktitle={International Conference on Machine Learning},
year={2024},
}

@article{Akner2024,
  title={Critique-out-loud reward models},
  author={Ankner, Zachary and Paul, Mansheej and Cui, Brandon and Chang, Jonathan D and Ammanabrolu, Prithviraj},
  journal={arXiv preprint arXiv:2408.11791},
  year={2024}
}

@inproceedings{Zhang2025,
title={Generative Verifiers: Reward Modeling as Next-Token Prediction},
author={Lunjun Zhang and Arian Hosseini and Hritik Bansal and Mehran Kazemi and Aviral Kumar and Rishabh Agarwal},
booktitle={International Conference on Learning Representations},
year={2025},
}

@inproceedings{Lambert2025,
  title={Rewardbench: Evaluating reward models for language modeling},
  author={Lambert, Nathan and Pyatkin, Valentina and Morrison, Jacob and Miranda, Lester James Validad and Lin, Bill Yuchen and Chandu, Khyathi and Dziri, Nouha and Kumar, Sachin and Zick, Tom and Choi, Yejin and others},
  booktitle={Findings of the Association for Computational Linguistics: NAACL 2025},
  pages={1755--1797},
  year={2025}
}

@inproceedings{Stienon2020,
  author = {Nisan Stiennon and Long Ouyang and Jeff Wu and Daniel M. Ziegler and Ryan Lowe and Chelsea Voss and Alec Radford and Dario Amodei and Paul Christiano},
  title = {Learning to summarize from human feedback},
  booktitle = {NeurIPS},
  year = 2020,
}

@article{Ganguli2022,
  title={Red teaming language models to reduce harms: Methods, scaling behaviors, and lessons learned},
  author={Ganguli, Deep and Lovitt, Liane and Kernion, Jackson and Askell, Amanda and Bai, Yuntao and Kadavath, Saurav and Mann, Ben and Perez, Ethan and Schiefer, Nicholas and Ndousse, Kamal and others},
  journal={arXiv preprint arXiv:2209.07858},
  year={2022}
}

@inproceedings{Wan2018,
author = {Wan, Mengting and McAuley, Julian},
title = {Item recommendation on monotonic behavior chains},
year = {2018},
booktitle = {Proceedings of ACM Conference on Recommender Systems},
pages = {86–94}
}

@inproceedings{Wan2019,
    title = "Fine-Grained Spoiler Detection from Large-Scale Review Corpora",
    author = "Wan, Mengting  and
      Misra, Rishabh  and
      Nakashole, Ndapa  and
      McAuley, Julian",
    booktitle = "Proceedings of Annual Meeting of the Association for Computational Linguistics",
    month = jul,
    year = "2019",
    pages = "2605--2610"
}

@inproceedings{Yu2025,
    title = "Self-Generated Critiques Boost Reward Modeling for Language Models",
    author = "Yu, Yue  and
      Chen, Zhengxing  and
      Zhang, Aston  and
      Tan, Liang  and
      Zhu, Chenguang  and
      Pang, Richard Yuanzhe  and
      Qian, Yundi  and
      Wang, Xuewei  and
      Gururangan, Suchin  and
      Zhang, Chao  and
      Kambadur, Melanie  and
      Mahajan, Dhruv  and
      Hou, Rui",
    booktitle = "Proceedings of Conference of the Nations of the Americas Chapter of the Association for Computational Linguistics: Human Language Technologies (Volume 1: Long Papers)",
    month = apr,
    year = "2025",
    pages = "11499--11514",
}

@inproceedings{Zheng2023,
title={Judging {LLM}-as-a-Judge with {MT}-Bench and Chatbot Arena},
author={Lianmin Zheng and Wei-Lin Chiang and Ying Sheng and Siyuan Zhuang and Zhanghao Wu and Yonghao Zhuang and Zi Lin and Zhuohan Li and Dacheng Li and Eric Xing and Hao Zhang and Joseph E. Gonzalez and Ion Stoica},
booktitle={Conference on Neural Information Processing Systems Datasets and Benchmarks Track},
year={2023}
}

@article{Clark2021,
  title={All that's' human'is not gold: Evaluating human evaluation of generated text},
  author={Clark, Elizabeth and August, Tal and Serrano, Sofia and Haduong, Nikita and Gururangan, Suchin and Smith, Noah A},
  journal={arXiv preprint arXiv:2107.00061},
  year={2021}
}

@inproceedings{Liu2023geval,
title={G-Eval: {NLG} Evaluation using Gpt-4 with Better Human Alignment},
author={Yang Liu and Dan Iter and Yichong Xu and Shuohang Wang and Ruochen Xu and Chenguang Zhu},
booktitle={Conference on Empirical Methods in Natural Language Processing},
year={2023}
}

@inproceedings{Kim2024,
title={Prometheus: Inducing Fine-Grained Evaluation Capability in Language Models},
author={Seungone Kim and Jamin Shin and Yejin Cho and Joel Jang and Shayne Longpre and Hwaran Lee and Sangdoo Yun and Seongjin Shin and Sungdong Kim and James Thorne and Minjoon Seo},
booktitle={International Conference on Learning Representations},
year={2024}
}

@inproceedings{Dubois2023,
title={AlpacaFarm: A Simulation Framework for Methods that Learn from Human Feedback},
author={Yann Dubois and Xuechen Li and Rohan Taori and Tianyi Zhang and Ishaan Gulrajani and Jimmy Ba and Carlos Guestrin and Percy Liang and Tatsunori Hashimoto},
booktitle={Conference on Neural Information Processing Systems},
year={2023}
}

@inproceedings{Saha2024,
  title={Branch-solve-merge improves large language model evaluation and generation},
  author={Saha, Swarnadeep and Levy, Omer and Celikyilmaz, Asli and Bansal, Mohit and Weston, Jason and Li, Xian},
  booktitle={Proceedings of Conference of the North American Chapter of the Association for Computational Linguistics: Human Language Technologies (Volume 1: Long Papers)},
  pages={8352--8370},
  year={2024}
}

@article{Wang2024self,
  title={Self-taught evaluators},
  author={Wang, Tianlu and Kulikov, Ilia and Golovneva, Olga and Yu, Ping and Yuan, Weizhe and Dwivedi-Yu, Jane and Pang, Richard Yuanzhe and Fazel-Zarandi, Maryam and Weston, Jason and Li, Xian},
  journal={arXiv preprint arXiv:2408.02666},
  year={2024}
}

@article{Trivedi2024,
  title={Self-rationalization improves llm as a fine-grained judge},
  author={Trivedi, Prapti and Gulati, Aditya and Molenschot, Oliver and Rajeev, Meghana Arakkal and Ramamurthy, Rajkumar and Stevens, Keith and Chaudhery, Tanveesh Singh and Jambholkar, Jahnavi and Zou, James and Rajani, Nazneen},
  journal={arXiv preprint arXiv:2410.05495},
  year={2024}
}

@article{Li2025scoring,
  title={Evaluating Scoring Bias in LLM-as-a-Judge},
  author={Li, Qingquan and Dou, Shaoyu and Shao, Kailai and Chen, Chao and Hu, Haixiang},
  journal={arXiv preprint arXiv:2506.22316},
  year={2025}
}

@article{Zhang2024reference,
  title={Reviseval: Improving llm-as-a-judge via response-adapted references},
  author={Zhang, Qiyuan and Wang, Yufei and Yu, Tiezheng and Jiang, Yuxin and Wu, Chuhan and Li, Liangyou and Wang, Yasheng and Jiang, Xin and Shang, Lifeng and Tang, Ruiming and others},
  journal={arXiv preprint arXiv:2410.05193},
  year={2024}
}

@inproceedings{Dsouza2025,
    title = "{YES}ci{E}val: Robust {LLM}-as-a-Judge for Scientific Question Answering",
    author = {D{'}Souza, Jennifer  and
      Babaei Giglou, Hamed  and
      M{\"u}nch, Quentin},
    booktitle = "Proceedings of Annual Meeting of the Association for Computational Linguistics (Volume 1: Long Papers)",
    month = jul,
    year = "2025",
    pages = "13749--13783"
}

@inproceedings{Saha2025,
title={Learning to Plan \& Reason for Evaluation with Thinking-{LLM}-as-a-Judge},
author={Swarnadeep Saha and Xian Li and Marjan Ghazvininejad and Jason E Weston and Tianlu Wang},
booktitle={International Conference on Machine Learning},
year={2025}
}

@inproceedings{Chen2021faithful,
    title = "Improving Faithfulness in Abstractive Summarization with Contrast Candidate Generation and Selection",
    author = "Chen, Sihao  and
      Zhang, Fan  and
      Sone, Kazoo  and
      Roth, Dan",
    booktitle = "Conference of the North American Chapter of the Association for Computational Linguistics: Human Language Technologies",
    month = jun,
    year = "2021",
    pages = "5935--5941"
}

@article{Hou2024chatglm,
  title={Chatglm-rlhf: Practices of aligning large language models with human feedback},
  author={Hou, Zhenyu and Niu, Yilin and Du, Zhengxiao and Zhang, Xiaohan and Liu, Xiao and Zeng, Aohan and Zheng, Qinkai and Huang, Minlie and Wang, Hongning and Tang, Jie and others},
  journal={arXiv preprint arXiv:2404.00934},
  year={2024}
}

@article{Rame2023,
  title={Rewarded soups: towards pareto-optimal alignment by interpolating weights fine-tuned on diverse rewards},
  author={Rame, Alexandre and Couairon, Guillaume and Dancette, Corentin and Gaya, Jean-Baptiste and Shukor, Mustafa and Soulier, Laure and Cord, Matthieu},
  journal={Advances in Neural Information Processing Systems},
  volume={36},
  pages={71095--71134},
  year={2023}
}

@article{Ouyang2022,
  title={Training language models to follow instructions with human feedback},
  author={Ouyang, Long and Wu, Jeffrey and Jiang, Xu and Almeida, Diogo and Wainwright, Carroll and Mishkin, Pamela and Zhang, Chong and Agarwal, Sandhini and Slama, Katarina and Ray, Alex and others},
  journal={Advances in neural information processing systems},
  volume={35},
  pages={27730--27744},
  year={2022}
}

@article{Chen2025rmr1,
  title={Rm-r1: Reward modeling as reasoning},
  author={Chen, Xiusi and Li, Gaotang and Wang, Ziqi and Jin, Bowen and Qian, Cheng and Wang, Yu and Wang, Hongru and Zhang, Yu and Zhang, Denghui and Zhang, Tong and others},
  journal={arXiv preprint arXiv:2505.02387},
  year={2025}
}

@inproceedings{Xie2025teaching,
title={Teaching Language Models to Critique via Reinforcement Learning},
author={Zhihui Xie and Jie chen and Liyu Chen and Weichao Mao and Jingjing Xu and Lingpeng Kong},
booktitle={International Conference on Machine Learning},
year={2025}
}

@article{Wei2022chain,
  title={Chain-of-thought prompting elicits reasoning in large language models},
  author={Wei, Jason and Wang, Xuezhi and Schuurmans, Dale and Bosma, Maarten and Xia, Fei and Chi, Ed and Le, Quoc V and Zhou, Denny and others},
  journal={Advances in neural information processing systems},
  volume={35},
  pages={24824--24837},
  year={2022}
}

@article{Belcak2025small,
  title={Small Language Models are the Future of Agentic AI},
  author={Belcak, Peter and Heinrich, Greg and Diao, Shizhe and Fu, Yonggan and Dong, Xin and Muralidharan, Saurav and Lin, Yingyan Celine and Molchanov, Pavlo},
  journal={arXiv preprint arXiv:2506.02153},
  year={2025}
}

@inproceedings{madaan2023self,
  title={Self-Refine: Iterative Refinement with Self-Feedback},
  author={Madaan, Aman and Tandon, Niket and Gupta, Prakhar and Hallinan, Skyler and Gao, Luyu and Wiegreffe, Sarah and Alon, Uri and Dziri, Nouha and Prabhumoye, Shrimai and Yang, Yiming and others},
  booktitle={Advances in Neural Information Processing Systems},
  volume={36},
  year={2023}
}

@inproceedings{hsieh2023distilling,
  title={Distilling step-by-step! outperforming larger language models with less training data and smaller model sizes},
  author={Hsieh, Cheng-Yu and Li, Chun-Liang and Yeh, Chih-Kuan and Nakhost, Hootan and Fujii, Yasuhisa and Ratner, Alex and Krishna, Ranjay and Lee, Chen-Yu and Pfister, Tomas},
  booktitle={Findings of the Association for Computational Linguistics: ACL 2023},
  pages={8003--8017},
  year={2023}
}

@inproceedings{magister2023teaching,
  title={Teaching small language models to reason},
  author={Magister, Lucie Charlotte and Mallinson, Jonathan and Adamek, Jakub and Malmi, Eric and Severyn, Aliaksei},
  booktitle={Proceedings of the 61st annual meeting of the association for computational linguistics (volume 2: short papers)},
  pages={1773--1781},
  year={2023}
}

@inproceedings{ho2023large,
  title={Large language models are reasoning teachers},
  author={Ho, Namgyu and Schmid, Laura and Yun, Se-Young},
  booktitle={Proceedings of the 61st annual meeting of the association for computational linguistics (volume 1: long papers)},
  pages={14852--14882},
  year={2023}
}

@article{hinton2015distilling,
  title={Distilling the knowledge in a neural network},
  author={Hinton, Geoffrey and Vinyals, Oriol and Dean, Jeff},
  journal={arXiv preprint arXiv:1503.02531},
  year={2015}
}

@article{sanh2019distilbert,
  title={DistilBERT, a distilled version of BERT: smaller, faster, cheaper and lighter},
  author={Sanh, Victor and Debut, Lysandre and Chaumond, Julien and Wolf, Thomas},
  journal={arXiv preprint arXiv:1910.01108},
  year={2019}
}

@article{littlestone1994weighted,
  title={The Weighted Majority Algorithm},
  author={Littlestone, Nick and Warmuth, Manfred K},
  journal={Information and Computation},
  volume={108},
  number={2},
  pages={212--261},
  year={1994}
}

@article{freund1997decision,
  title={A Decision-Theoretic Generalization of On-Line Learning and an Application to Boosting},
  author={Freund, Yoav and Schapire, Robert E},
  journal={Journal of Computer and System Sciences},
  volume={55},
  number={1},
  pages={119--139},
  year={1997}
}

@book{cesa2006prediction,
  title={Prediction, Learning, and Games},
  author={Cesa-Bianchi, Nicol{\`o} and Lugosi, G{\'a}bor},
  year={2006},
  publisher={Cambridge University Press}
}

@article{cesa2006worst,
  title={Worst-Case Analysis of Selective Sampling for Linear Classification},
  author={Cesa-Bianchi, Nicol{\`o} and Gentile, Claudio and Zaniboni, Luca},
  journal={Journal of Machine Learning Research},
  volume={7},
  pages={1205--1230},
  year={2006}
}

@article{dekel2012selective,
  title={Selective Sampling and Active Learning from Single and Multiple Teachers},
  author={Dekel, Ofer and Gentile, Claudio and Sridharan, Karthik},
  journal={Journal of Machine Learning Research},
  volume={13},
  pages={2655--2697},
  year={2012}
}

@inproceedings{agarwal2017corralling,
  title={Corralling a Band of Bandit Algorithms},
  author={Agarwal, Alekh and Luo, Haipeng and Neyshabur, Behnam and Schapire, Robert E},
  booktitle={Proceedings of the 30th Conference on Learning Theory (COLT)},
  pages={12--38},
  year={2017}
}

@article{foster2020instance,
  title={Instance-dependent complexity of contextual bandits and reinforcement learning: A disagreement-based perspective},
  author={Foster, Dylan J and Rakhlin, Alexander and Simchi-Levi, David and Xu, Yunzong},
  journal={arXiv preprint arXiv:2010.03104},
  year={2020}
}

@techreport{settles2009active,
  title={Active Learning Literature Survey},
  author={Settles, Burr},
  year={2009},
  institution={University of Wisconsin--Madison},
  number={1648}
}

@article{openai2025gptoss120b,
  title={gpt-oss-120b \& gpt-oss-20b model card},
  author={Agarwal, Sandhini and Ahmad, Lama and Ai, Jason and Altman, Sam and Applebaum, Andy and Arbus, Edwin and Arora, Rahul K and Bai, Yu and Baker, Bowen and Bao, Haiming and others},
  journal={arXiv preprint arXiv:2508.10925},
  year={2025}
}

@article{gemma2024,
  title={Gemma: Open models based on gemini research and technology},
  author={Team, Gemma and Mesnard, Thomas and Hardin, Cassidy and Dadashi, Robert and Bhupatiraju, Surya and Pathak, Shreya and Sifre, Laurent and Rivi{\`e}re, Morgane and Kale, Mihir Sanjay and Love, Juliette and others},
  journal={arXiv preprint arXiv:2403.08295},
  year={2024}
}

@article{qwen3technicalreport,
  title={Qwen3 technical report},
  author={Yang, An and Li, Anfeng and Yang, Baosong and Zhang, Beichen and Hui, Binyuan and Zheng, Bo and Yu, Bowen and Gao, Chang and Huang, Chengen and Lv, Chenxu and others},
  journal={arXiv preprint arXiv:2505.09388},
  year={2025}
}

@inproceedings{Hu2022lora,
title={Lo{RA}: Low-Rank Adaptation of Large Language Models},
author={Edward J Hu and yelong shen and Phillip Wallis and Zeyuan Allen-Zhu and Yuanzhi Li and Shean Wang and Lu Wang and Weizhu Chen},
booktitle={International Conference on Learning Representations},
year={2022}
}

@article{Schulman2017proximal,
  title={Proximal policy optimization algorithms},
  author={Schulman, John and Wolski, Filip and Dhariwal, Prafulla and Radford, Alec and Klimov, Oleg},
  journal={arXiv preprint arXiv:1707.06347},
  year={2017}
}

@article{Shao2024deepseekmath,
  title={Deepseekmath: Pushing the limits of mathematical reasoning in open language models},
  author={Shao, Zhihong and Wang, Peiyi and Zhu, Qihao and Xu, Runxin and Song, Junxiao and Bi, Xiao and Zhang, Haowei and Zhang, Mingchuan and Li, YK and Wu, Yang and others},
  journal={arXiv preprint arXiv:2402.03300},
  year={2024}
}

@article{Shen2019raker,
  author  = {Yanning Shen and Tianyi Chen and Georgios B. Giannakis},
  title   = {Random Feature-based Online Multi-kernel Learning in Environments with Unknown Dynamics},
  journal = {Journal of Machine Learning Research},
  year    = {2019},
  volume  = {20},
  number  = {22},
  pages   = {1--36}
}

@InProceedings{Alon2015,
  title = 	 {Online Learning with Feedback Graphs: Beyond Bandits},
  author = 	 {Alon, Noga and Cesa-Bianchi, Nicolò and Dekel, Ofer and Koren, Tomer},
  booktitle = 	 {Proceedings of Conference on Learning Theory},
  pages = 	 {23--35},
  year = 	 {2015},
  volume = 	 {40},
  month = 	 {03--06 Jul},
}

@inproceedings{deepspeed,
author = {Rasley, Jeff and Rajbhandari, Samyam and Ruwase, Olatunji and He, Yuxiong},
title = {DeepSpeed: System Optimizations Enable Training Deep Learning Models with Over 100 Billion Parameters},
year = {2020},
booktitle = {Proceedings of ACM SIGKDD International Conference on Knowledge Discovery \& Data Mining},
pages = {3505–3506}
}

@article{Hartmann2025,
title = {The power of generative marketing: Can generative AI create superhuman visual marketing content?},
journal = {International Journal of Research in Marketing},
volume = {42},
number = {1},
pages = {13-31},
year = {2025},
author = {Jochen Hartmann and Yannick Exner and Samuel Domdey}
}

@article{Dorka2024,
  title={Quantile regression for distributional reward models in rlhf},
  author={Dorka, Nicolai},
  journal={arXiv preprint arXiv:2409.10164},
  year={2024}
}

@article{Liu2024skywork,
  title={Skywork-Reward: Bag of Tricks for Reward Modeling in LLMs},
  author={Liu, Chris Yuhao and Zeng, Liang and Liu, Jiacai and Yan, Rui and He, Jujie and Wang, Chaojie and Yan, Shuicheng and Liu, Yang and Zhou, Yahui},
  journal={arXiv preprint arXiv:2410.18451},
  year={2024}
}

@inproceedings{helpsteer3,
    title = "{H}elp{S}teer3: Human-Annotated Feedback and Edit Data to Empower Inference-Time Scaling in Open-Ended General-Domain Tasks",
    author = "Wang, Zhilin  and
      Zeng, Jiaqi  and
      Delalleau, Olivier  and
      Egert, Daniel  and
      Evans, Ellie  and
      Shin, Hoo-Chang  and
      Soares, Felipe  and
      Dong, Yi  and
      Kuchaiev, Oleksii",
    booktitle = "Proceedings of the Annual Meeting of the Association for Computational Linguistics (Volume 1: Long Papers)",
    month = jul,
    year = "2025",
    publisher = "Association for Computational Linguistics",
    pages = "25640--25662",
}

@article{Xu2026alternating,
  title={Alternating reinforcement learning for rubric-based reward modeling in non-verifiable llm post-training},
  author={Xu, Ran and Liu, Tianci and Dong, Zihan and Yu, Tony and Hong, Ilgee and Yang, Carl and Zhang, Linjun and Zhao, Tao and Wang, Haoyu},
  journal={arXiv preprint arXiv:2602.01511},
  year={2026}
}

@inproceedings{Kawabata2026,
  title={C2: Scalable rubric-augmented reward modeling from binary preferences},
  author={Kawabata, Akira and Sugawara, Saku},
  booktitle={Proceedings of the Annual Meeting of the Association for Computational Linguistics (Volume 1: Long Papers)},
  pages={11397--11424},
  year={2026}
}

@inproceedings{
malik2026rewardbench,
title={RewardBench 2: Advancing Reward Model Evaluation},
author={Saumya Malik and Valentina Pyatkin and Sander Land and Jacob Morrison and Noah A. Smith and Hannaneh Hajishirzi and Nathan Lambert},
booktitle={International Conference on Learning Representations},
year={2026}
}

@article{gemma2,
  title={Gemma 2: Improving open language models at a practical size},
  author={Team, Gemma and Riviere, Morgane and Pathak, Shreya and Sessa, Pier Giuseppe and Hardin, Cassidy and Bhupatiraju, Surya and Hussenot, L{\'e}onard and Mesnard, Thomas and Shahriari, Bobak and Ram{\'e}, Alexandre and others},
  journal={arXiv preprint arXiv:2408.00118},
  year={2024}
}

@article{gemma3,
  title={Gemma 3 technical report},
  author={Team, Gemma and Kamath, Aishwarya and Ferret, Johan and Pathak, Shreya and Vieillard, Nino and Merhej, Ramona and Perrin, Sarah and Matejovicova, Tatiana and Ram{\'e}, Alexandre and Rivi{\`e}re, Morgane and others},
  journal={arXiv preprint arXiv:2503.19786},
  year={2025}
}

\newpage
\appendix
\onecolumn

\section{Analysis of the RLHF Reward Function} \label{sec:analysis-reward}
The following theorem establishes the relationship between the reward function in \eqref{eq:4} and the Kullback--Leibler (KL) divergence between the true distribution $p$ and the rating policy $\pi_\vtheta$ in the presence of generated explanations.
\begin{theorem} \label{th:2}
The expected reward in \eqref{eq:4} is upper-bounded by
\begin{align}
\E_{p, \pi_{\boldsymbol{\rho}}}[R(\ve|\vx, \vy)] \le -D_{\text{KL}}(p(\vx, \vy, r)||\pi_\vtheta(\vx, \vy, r)) + \E_p[\log(p(r|\vx, \vy))]
\end{align}
where the marginal distribution $\pi_\vtheta(\vx, \vy, r)$ is defined as $\pi_\vtheta(\vx, \vy, r) = \sum_{\ve}{\pi_\vtheta(\vx, \vy, \ve, r)}$ and $\ve$ denotes an explanation.
\end{theorem}

The proof can be found in Appendix \ref{proof:th2}. Theorem \ref{th:2} shows that the expected reward is bounded from above by the negative KL divergence between $p$ and $\pi_\vtheta$. It is noteworthy that the term $\E_p[\log(p(r|\vx, \vy))]$ is independent of the learning parameters. On the other hand, the joint distribution $\pi_\vtheta(\vx, \vy, \ve, r)$ depends on the generated explanation $\ve$. Therefore, changes in $\pi_{\boldsymbol{\rho}}$ can induce changes in $\pi_\vtheta(\vx, \vy, \ve, r)$.
Theorem \ref{th:2} indicates that a smaller KL divergence between $p$ and $\pi_\vtheta$ corresponds to a larger upper bound on the expected reward. Since the true distribution $p$ is unknown and therefore cannot be directly used to minimize the KL divergence, the reward $R$ defined in \eqref{eq:4} provides a practical surrogate objective. Maximizing the reward encourages $\pi_\vtheta$ to approach the true distribution $p$ by reducing their KL divergence.

\section{Online Ensemble Learning} \label{sec:ens-sm}
Here, we present an online ensemble learning algorithm that dynamically combines the predictions of the teacher model with those of the rating module.

Let $\sD_T := \{(\vx_i, \vy_i)\}$ denote a test set of prompt–response pairs, where the task is to rate the quality of each response $\vy_i$ given the corresponding prompt $\vx_i$. At each time step $t$, a prompt–response pair $(\vx_t, \vy_t)$ is sampled from the dataset. There are two options for performing the evaluation. One option is to employ the trained policies $\pi_{\boldsymbol{\rho}}$ and $\pi_\vtheta$. The other option is to employ an ensemble of the trained policies $\pi_{\boldsymbol{\rho}}$ and $\pi_\vtheta$, combined with the teacher’s rating of response quality. Let $\hat{r}_{m,t}$ denote the rating predicted by the rating policy $\pi_\vtheta$ for $(\vx_t, \vy_t)$, and let $\hat{r}_{a,t}$ denote the rating provided by the teacher. The ensemble rating is constructed as follows:
\begin{align} \label{eq:ens_rating}
    \hat{r}_{e,t} = \boldsymbol{\omega}_t^\top \hat{\vr}_t,
\end{align}
where $\boldsymbol{\omega}_t = (\omega_{a,t}, \omega_{m,t})$ is a learnable weight vector such that $\omega_{a,t}$ and $\omega_{m,t}$ are assigned to the teacher and the rating policy $\pi_\vtheta$, respectively. Moreover, $\hat{\vr}_t$ denotes the vector $\hat{\vr}_t := (\hat{r}_{a,t}, \hat{r}_{m,t})$. In order to assess the ratings given by the teacher and rating policy $\pi_\vtheta$, a ground truth rating is required. Here, we assume that there is access to labelers such as human annotators who can provide their rating $r_t$ for $(\vx_t, \vy_t)$.  However, since querying for a human label might be costly, for a time horizon $T$ a budget $S$ for label querying can be set. To not violate the label querying budget in expectation, at each time step, the label is queried with probability $s = \frac{S}{T}$. Moreover, calling the teacher for rating may add computational overhead and latency. Therefore, at time step $t$, the teacher is called with probability
\begin{align}
    \psi_t = (1-\epsilon)\frac{w_{e,t}}{w_{m,t}+w_{e,t}} + \frac{\epsilon}{2}, \label{eq:9}
\end{align}
where $\epsilon \ge 0$ is an exploration coefficient while $w_{e,t}$ and $w_{m,t}$ are learnable weights assigned to ensemble rating and rating policy $\pi_\vtheta$, respectively to indicate their credibility. According to \eqref{eq:9}, the probability of calling the teacher increases with its weight $w_{e,t}$. When the teacher is not called, the rating task is done solely by the rating policy $\pi_\vtheta$. If the label is queried at time step $t$, the loss of rating prediction $\hat{r}_{e,t}$ can be computed as $L(\hat{r}_{e,t}, r_t)$ where $L(\cdot,\cdot)$ is a loss function such as mean absolute error (MAE). Since the ground truth label $r_t$ may not be available at every time step, the importance sampling estimate of loss is used. Let $\hat{\ell}_{e,t}$ and $\hat{\ell}_{m,t}$ denote the importance sampling loss estimates for ratings given by the ensemble and the rating policy $\pi_{\vtheta}$, respectively which can be obtained as:
\begin{align} \label{eq:10}
    \hat{\ell}_{e,t} = \frac{L(\hat{r}_{e,t}, r_t)}{s\psi_t} \1_l \1_a, \quad
    \hat{\ell}_{m,t} = \frac{L(\hat{r}_{m,t}, r_t)}{s} \1_l,
\end{align}
where $\1_a$ and $\1_l$ denote indicator functions. The indicator function $\1_a$ equals $1$ if the teacher is queried to provide a rating, and $0$ otherwise. Similarly, $\1_l$ equals $1$ if the label is queried, and $0$ otherwise. Note that the expected values of the importance-sampling loss estimates $\hat{\ell}_{e,t}$ and $\hat{\ell}_{m,t}$ correspond to the true losses $L(\hat{r}_{e,t}, r_t)$ and $L(\hat{r}_{m,t}, r_t)$, respectively.
Using the multiplicative update rule, the weights $w_{e,t}$ and $w_{m,t}$ are updated as:
\begin{align} \label{eq:weight_update}
    w_{e,t+1} = w_{e,t}\exp(-\eta \hat{\ell}_{e,t}), \quad
    w_{m,t+1} = w_{m,t}\exp(-\eta \hat{\ell}_{m,t}).
\end{align}
Furthermore, the weight vector $\boldsymbol{\omega}_t$ is updated using gradient descent as follows:
\begin{align}
    \boldsymbol{\omega}_{t+1} = \boldsymbol{\omega}_t - \eta \nabla \hat{\ell}_{e,t}. \label{eq:12}
\end{align}
The regret of the proposed online ensemble learning approach is analyzed in Theorem \ref{th:1}.
\begin{theorem} \label{th:1}
Assume that the loss function $L(\cdot,\cdot)$ is convex with respect to weight $\boldsymbol{\omega}$, and that the loss and its gradient are bounded such that
$0 \le L(r_1, r_2) \le 1$ and $\|\nabla L(r_1, r_2)\| \le G$.
Let $\tilde{r}_{e,t}$ denote the final rating produced by the proposed algorithm.
Moreover, let $\boldsymbol{\omega}^*$ denote the optimal parameter defined as
\begin{align}
\boldsymbol{\omega}^* = \argmin_{\boldsymbol{\omega}} \sum_{t=1}^{T}{L(r_{e,t}, r_t)},
\quad r_{e,t} = \boldsymbol{\omega}^\top \hat{\vr}_t .
\end{align}
Then, the expected regret of the proposed algorithm with respect to the best ensemble model is bounded from above by
\begin{align}
& \sum_{t=1}^{T} \E\!\left[L(\tilde{r}_{e,t}, r_t)\right]
- \sum_{t=1}^{T} L(r^*_{e,t}, r_t)
\le\;
\frac{2\ln 2 + \|\boldsymbol{\omega}^*\|^2}{2\eta}
+ \left(\epsilon
+ \frac{\eta}{s\epsilon}(1+\epsilon+G^2)\right) T ,
\label{reg:bound}
\end{align}
where $r^*_{e,t} = \boldsymbol{\omega}^{*\top} \hat{\vr}_t$.
\end{theorem}
The proof of Theorem \ref{th:1} is provided in Appendix \ref{proof:th1}. According to \eqref{reg:bound}, setting $\eta = \gO(T^{-\frac{2}{3}})$ and $\epsilon = \gO(T^{-\frac{1}{3}})$, the proposed ensemble approach achieves sublinear regret of $\gO(T^{\frac{2}{3}})$.

\section{Proofs}
\subsection{Proof of Theorem \ref{th:2}} \label{proof:th2}
\begin{proof}
The negative KL divergence between the true distribution $p$ and the policy $\pi_\vtheta$ can be expressed as follows:
\begin{align}
    - D_{\text{KL}}(p(\vx, \vy, r)||\pi_\vtheta(\vx, \vy, r)) =& \sum_{(\vx, \vy, r)}p(\vx, \vy, r)\log\left(\sum_{\ve}\pi_\vtheta(\vx, \vy, \ve, r)\right) \nonumber \\ & - \sum_{(\vx, \vy, r)}{p(\vx, \vy, r)\log(p(\vx, \vy, r))}.
\end{align}
The above equation can be rewritten as:
\begin{align}
    - D_{\text{KL}}(p||\pi_\vtheta) =& \sum_{(\vx, \vy, r)}{p(\vx, \vy, r)\log\left(p(\vx, \vy)\sum_{\ve}\pi_{\boldsymbol{\rho}}(\ve|\vx, \vy)\pi_\vtheta(r|\vx, \vy, \ve)\right)} \nonumber \\ &- \sum_{(\vx, \vy, r)}{p(\vx, \vy, r)\log(p(\vx, \vy, r))}. \label{eq:17ap}
\end{align}
Since $\log(x)$ is concave for $x>0$, and $\pi_{\boldsymbol{\rho}}(\ve|\vx,\vy)$ defines a probability distribution over $\ve$, Jensen's inequality gives
\begin{align}
    \log\left(p(\vx, \vy)\sum_{\ve}{\pi_{\boldsymbol{\rho}}(\ve|\vx, \vy)\pi_\vtheta(r|\vx, \vy, \ve)}\right) \ge \sum_{\ve}{\pi_{\boldsymbol{\rho}}(\ve|\vx, \vy)\log(\pi_\vtheta(r|\vx, \vy, \ve))} + \log(p(\vx, \vy)). \label{eq:18ap}
\end{align}
Substituting \eqref{eq:18ap} into \eqref{eq:17ap}, we get
\begin{align}
    - D_{\text{KL}}(p||\pi_\vtheta) \ge & \sum_{(\vx, \vy, r)}{\sum_{\ve}{p(\vx, \vy, r)\pi_{\boldsymbol{\rho}}(\ve|\vx, \vy)\log(\pi_\vtheta(r|\vx, \vy, \ve))}} \nonumber \\ &+ \sum_{(\vx, \vy)}{p(\vx, \vy)\log(p(\vx, \vy))} - \sum_{(\vx, \vy, r)}{p(\vx, \vy, r)\log(p(\vx, \vy, r))}. \label{eq:22ap}
\end{align}
Let $p(\vx, \vy, \ve, r) := p(\vx, \vy, r)\pi_{\boldsymbol{\rho}}(\ve|\vx, \vy)$ denote the true probability of observing $(\vx, \vy, \ve, r)$. It is noteworthy that the expected value of $\log(\pi_\vtheta(r|\vx, \vy, \ve))$ under the true probability distribution $p$ can be expressed as:
\begin{align}
\E_{p,\pi_{\boldsymbol{\rho}}}[\log(\pi_\vtheta(r|\vx, \vy, \ve))] = \sum_{(\vx, \vy,\ve, r)}{p(\vx, \vy, r)\pi_{\boldsymbol{\rho}}(\ve|\vx, \vy)\log(\pi_\vtheta(r|\vx, \vy, \ve))}. \label{eq:19ap}
\end{align}
Furthermore, since $p(r|\vx, \vy) = \frac{p(\vx, \vy, r)}{p(\vx, \vy)}$, we can write that
\begin{align}
    - \E_p[\log(p(r|\vx, \vy))] = \sum_{(\vx, \vy)}{p(\vx, \vy)\log(p(\vx, \vy))} - \sum_{(\vx, \vy, r)}{p(\vx, \vy, r)\log(p(\vx, \vy, r))}. \label{eq:20ap}
\end{align}
Substituting \eqref{eq:19ap} and \eqref{eq:20ap} into \eqref{eq:22ap}, we obtain
\begin{align}
-D_{\text{KL}}(p||\pi_\vtheta) \ge \E_{p,\pi_{\boldsymbol{\rho}}}[\log(\pi_\vtheta(r|\vx, \vy, \ve))] - \E_p[\log(p(r|\vx, \vy))], \label{eq:21ap}
\end{align}
which proves the Theorem.
\end{proof}

\subsection{Proof of Theorem \ref{th:1}} \label{proof:th1}
\begin{proof}
Let $W_t$ be defined as $W_t := w_{e,t} + w_{m,t}$. Then, we can write
\begin{align}
    \frac{W_{t+1}}{W_t} = \frac{w_{e,t}}{W_t}\exp(-\eta \hat{\ell}_{e,t}) + \frac{w_{m,t}}{W_t}\exp(-\eta \hat{\ell}_{m,t}). \label{eq:1ap}
\end{align}
According to \eqref{eq:9}, \eqref{eq:1ap} can be rewritten as:
\begin{align}
    \frac{W_{t+1}}{W_t} = \frac{\psi_t - \frac{\epsilon}{2}}{1-\epsilon}\exp(-\eta \hat{\ell}_{e,t}) + \left(1-\frac{\psi_t - \frac{\epsilon}{2}}{1-\epsilon}\right)\exp(-\eta \hat{\ell}_{m,t}).
\end{align}
Using the inequality $e^{-x} \le 1 - x + \frac{x^2}{2}$, $\forall x \ge 0$, we can conclude the following inequality:
\begin{align}
    \frac{W_{t+1}}{W_t} \le \frac{\psi_t - \frac{\epsilon}{2}}{1-\epsilon}(1-\eta \hat{\ell}_{e,t} + \frac{\eta^2}{2} \hat{\ell}_{e,t}^2) + \left(1-\frac{\psi_t - \frac{\epsilon}{2}}{1-\epsilon}\right)(1-\eta \hat{\ell}_{m,t} + \frac{\eta^2}{2} \hat{\ell}_{m,t}^2). \label{eq:3ap}
\end{align}
Taking logarithm from both sides of \eqref{eq:3ap} and considering the inequality $1+x \le e^x$, we arrive at:
\begin{align}
    \ln \frac{W_{t+1}}{W_t} \le \frac{\psi_t - \frac{\epsilon}{2}}{1-\epsilon}(-\eta \hat{\ell}_{e,t} + \frac{\eta^2}{2} \hat{\ell}_{e,t}^2) + \left(1-\frac{\psi_t - \frac{\epsilon}{2}}{1-\epsilon}\right)(-\eta \hat{\ell}_{m,t} + \frac{\eta^2}{2} \hat{\ell}_{m,t}^2). \label{eq:4ap}
\end{align}
Summing \eqref{eq:4ap} over time we get
\begin{align}
    \ln \frac{W_{T+1}}{W_1} \le \sum_{t=1}^{T}{\left(\frac{\psi_t - \frac{\epsilon}{2}}{1-\epsilon}(-\eta \hat{\ell}_{e,t} + \frac{\eta^2}{2} \hat{\ell}_{e,t}^2) + \left(1-\frac{\psi_t - \frac{\epsilon}{2}}{1-\epsilon}\right)(-\eta \hat{\ell}_{m,t} + \frac{\eta^2}{2} \hat{\ell}_{m,t}^2)\right)}. \label{eq:5ap}
\end{align}
Furthermore, $\ln \frac{W_{T+1}}{W_1}$ can be bounded from below as:
\begin{align}
    \ln \frac{W_{T+1}}{W_1} \ge \ln \frac{w_{e,T+1}}{W_1} = - \eta \sum_{t=1}^{T}{\hat{\ell}_{e,t}} - \ln 2 \label{eq:6ap}
\end{align}
Combining \eqref{eq:5ap} with \eqref{eq:6ap}, we get
\begin{align}
    & \sum_{t=1}^{T}{\frac{\eta \psi_t}{1-\epsilon} \hat{\ell}_{e,t} + \frac{\eta (1-\psi_t)}{1-\epsilon} \hat{\ell}_{m,t}} - \eta \sum_{t=1}^{T}{\hat{\ell}_{e,t}} \nonumber \\ \le & \ln 2 + \sum_{t=1}^{T}{\frac{\eta \epsilon}{2(1-\epsilon)}(\hat{\ell}_{e,t}+\hat{\ell}_{m,t})} + \frac{\eta^2}{2} \sum_{t=1}^{T}{\frac{\psi_t - \frac{\epsilon}{2}}{1-\epsilon}\hat{\ell}_{e,t}^2 + \frac{1- \psi_t - \frac{\epsilon}{2}}{1-\epsilon}\hat{\ell}_{m,t}^2} \label{eq:7ap}
\end{align}
Multiplying both sides of \eqref{eq:7ap} by $\frac{1-\epsilon}{\eta}$ and expanding $\hat{\ell}_{e,t}$ and $\hat{\ell}_{m,t}$ according to \eqref{eq:10}, we arrive at
\begin{align}
    & \sum_{t=1}^{T}{\psi_t\frac{L(\hat{r}_{e,t}, r_t)}{s\psi_t} \1_l \1_a + (1-\psi_t)\frac{L(\hat{r}_{m,t}, r_t)}{s} \1_l} - \sum_{t=1}^{T}{\frac{(1-\epsilon) L(\hat{r}_{e,t}, r_t)}{s\psi_t} \1_l \1_a } \nonumber \\ \le & \frac{\ln 2}{\eta} + \frac{\epsilon}{2} \sum_{t=1}^{T}{\frac{L(\hat{r}_{e,t}, r_t)}{s\psi_t} \1_l \1_a + \frac{L(\hat{r}_{m,t}, r_t)}{s} \1_l} + \frac{\eta}{2} \sum_{t=1}^{T}{\frac{L^2(\hat{r}_{e,t}, r_t)}{s^2\psi^2_t} \1_l \1_a + \frac{L^2(\hat{r}_{m,t}, r_t)}{s^2} \1_l} \label{eq:8ap}
\end{align}
Taking the expectation from both sides of \eqref{eq:8ap} with respect to $\1_a$ and $\1_l$ along with considering the assumptions that $0 \le L(\hat{r}_{e,t}, r_t) \le 1$ and $0 \le L(\hat{r}_{m,t}, r_t) \le 1$, we obtain
\begin{align}
    \sum_{t=1}^{T}{\psi_t L(\hat{r}_{e,t}, r_t) + (1-\psi_t)L(\hat{r}_{m,t}, r_t) } - \sum_{t=1}^{T}{L(\hat{r}_{e,t}, r_t)} \le \frac{\ln 2}{\eta} + \epsilon T + \frac{\eta}{2} \sum_{t=1}^{T}{\frac{1+\psi_t}{s\psi_t}} \label{eq:9ap}
\end{align}
Considering the fact that $\E[L(\tilde{r}_{e,t}, r_t)] = \psi_t L(\hat{r}_{e,t}, r_t) + (1-\psi_t)L(\hat{r}_{m,t}, r_t)$ along with using the convexity of the loss function $L(\cdot, \cdot)$ and Jensen's inequality, from \eqref{eq:9ap} we can conclude
\begin{align}
    \sum_{t=1}^{T}{\E[L(\tilde{r}_{e,t}, r_t)] } - \sum_{t=1}^{T}{L(\hat{r}_{e,t}, r_t)} \le \frac{\ln 2}{\eta} + \epsilon T + \frac{\eta}{2} \sum_{t=1}^{T}{\frac{1+\psi_t}{s\psi_t}}. \label{eq:10ap}
\end{align}
The regret of selecting between the ensemble model and relying on the rating tool, with respect to the ensemble model, is obtained in \eqref{eq:10ap}. The next step is to obtain the regret of the ensemble model with respect to the best one in hindsight. According to \eqref{eq:12}, for any $\boldsymbol{\omega}$ we can write:
\begin{align}
    \|\boldsymbol{\omega}_{t+1} - \boldsymbol{\omega}\|^2 = \|\boldsymbol{\omega}_t - \eta \nabla \hat{\ell}_{e,t} - \boldsymbol{\omega}\|^2 = \|\boldsymbol{\omega}_t - \boldsymbol{\omega}\|^2 - 2 \eta \nabla \hat{\ell}_{e,t}^\top (\boldsymbol{\omega}_t - \boldsymbol{\omega}) + \eta^2 \|\nabla \hat{\ell}_{e,t}\|^2.
\end{align}
Taking the expectation with respect to $\1_l$ and $\1_a$, we get
\begin{align}
\|\boldsymbol{\omega}_{t+1} - \boldsymbol{\omega}\|^2 = \|\boldsymbol{\omega}_t - \boldsymbol{\omega}\|^2 - 2 \eta \nabla L(\hat{r}_{e,t}, r_t)^\top (\boldsymbol{\omega}_t - \boldsymbol{\omega}) + \frac{\eta^2}{s\psi_t} \|\nabla L(\hat{r}_{e,t}, r_t)\|^2. \label{eq:12ap}
\end{align}
Taking the convexity of the loss function $L(\cdot,\cdot)$ into account, we can write
\begin{align}
    \nabla L(\hat{r}_{e,t}, r_t)^\top (\boldsymbol{\omega}_t - \boldsymbol{\omega}) \ge L(\hat{r}_{e,t}, r_t) - L(r_{e,t}, r_t) \label{eq:13ap}
\end{align}
where $r_{e,t}$ denotes the ensemble rating at time step $t$ corresponding to $\boldsymbol{\omega}$. Combining \eqref{eq:12ap} with \eqref{eq:13ap} and considering the assumption that $\|\nabla L(\hat{r}_{e,t}, r_t)\|^2 \le G^2$, we obtain
\begin{align}
    \|\boldsymbol{\omega}_{t+1} - \boldsymbol{\omega}\|^2 \le \|\boldsymbol{\omega}_t - \boldsymbol{\omega}\|^2 - 2 \eta \left(L(\hat{r}_{e,t}, r_t) - L(r_{e,t}, r_t)\right) + \frac{\eta^2 G^2}{s\psi_t}. \label{eq:14ap}
\end{align}
Summing \eqref{eq:14ap} over time and considering that $\boldsymbol{\omega}_1 = \boldsymbol{0}$, we get
\begin{align}
    \sum_{t=1}^{T}{L(\hat{r}_{e,t}, r_t)} - \sum_{t=1}^{T}{L(r_{e,t}, r_t)} \le \frac{\|\boldsymbol{\omega}\|^2 - \|\boldsymbol{\omega}_{T+1}-\boldsymbol{\omega}\|^2}{2\eta} + \sum_{t=1}^{T}{\frac{\eta G^2}{2s\psi_t}}. \label{eq:15ap}
\end{align}
Substituting $\boldsymbol{\omega}$ with $\boldsymbol{\omega}^*$ in \eqref{eq:15ap}, we can write
\begin{align}
    \sum_{t=1}^{T}{L(\hat{r}_{e,t}, r_t)} - \sum_{t=1}^{T}{L(r^*_{e,t}, r_t)} \le \frac{\|\boldsymbol{\omega}^*\|^2}{2\eta} + \sum_{t=1}^{T}{\frac{\eta G^2}{2s\psi_t}} \label{eq:16ap}
\end{align}
where $r^*_{e,t}$ denotes the ensemble rating at time step $t$ corresponding to $\boldsymbol{\omega}^*$. Combining \eqref{eq:16ap} with \eqref{eq:10ap} we arrive at
\begin{align}
    \sum_{t=1}^{T}{\E[L(\tilde{r}_{e,t}, r_t)] } - \sum_{t=1}^{T}{L(r^*_{e,t}, r_t)} \le \frac{2\ln 2 + \|\boldsymbol{\omega}^*\|^2}{2\eta} + \epsilon T + \frac{\eta}{2} \sum_{t=1}^{T}{\frac{1+\psi_t + G^2}{s\psi_t}}
\end{align}
Considering that $\psi_t \ge \frac{\epsilon}{2}$, we obtain
\begin{align}
    \sum_{t=1}^{T}{\E[L(\tilde{r}_{e,t}, r_t)] } - \sum_{t=1}^{T}{L(r^*_{e,t}, r_t)} \le \frac{2\ln 2 + \|\boldsymbol{\omega}^*\|^2}{2\eta} + \left(\epsilon+\frac{\eta}{s\epsilon}(1+\epsilon+G^2)\right)T
\end{align}
which proves the Theorem.
\end{proof}

\section{Supplementary Experimental Details and Results} \label{supp:exp}

\subsection{Datasets}
\textbf{Summarize from Feedback:} This dataset contains original texts paired with their corresponding summaries. The texts consist of either Reddit posts or news articles. Human annotators scored the accuracy of each summary using a scale from 1 to 7 \cite{Stienon2020}. To unify the scoring scales across datasets, we normalize each score by dividing it by 7, ensuring that all normalized scores lie in the interval $[0,1]$. For training and validation, we use samples based on Reddit posts, with 90\% of the samples used for training and the remaining 10\% for validation. To examine the generalizability of the models, samples based on news articles are used for testing.

\textbf{HH-RLHF:} This dataset contains conversations between humans and an AI assistant. The evaluation task measures the success of a human adversary in eliciting harmful instructions from the assistant, which is rated by human annotators \cite{Ganguli2022}. The original scores are on a scale from 0 to 4; to unify the scoring scales across datasets, we normalize the scores by dividing them by 4. The dataset is split into 70\% training, 10\% validation, and 20\% test.

\textbf{Goodreads:} This dataset consists of book reviews and ratings provided by Goodreads users \cite{Wan2018,Wan2019}. The task is to predict user satisfaction with a book in order to evaluate the performance of personalized recommender systems. Each user rates books on a scale from 1 to 5; accordingly, we normalize the scores by dividing them by 5. For training, we extract 50,000 samples, where each sample corresponds to a user–book pair. Each sample includes 10 review examples for the corresponding book and 10 review examples from the corresponding user, each accompanied by a rating. The task is to predict the user’s rating for a given book by extracting user preferences and book content from the reviews. For testing, we extract a set of user–book pairs such that neither the book nor the user appears in the training data. Furthermore, to ensure that samples do not exceed the maximum input length of the base models, we filter out samples longer than 8,192 tokens. After filtering, we retain 1,912 samples for testing.

\textbf{GenImageNet:} This dataset includes a total of 10,320 AI-generated images, each associated with a prompt and 7-point Likert ratings for the perceived quality of the images. Similar to the Summarize from Feedback dataset, we normalize each score by dividing it by 7, ensuring that all normalized scores lie in the interval $[0,1]$. We split the images into 80\% training, 10\% validation, and 10\% test sets such that the models are evaluated on unseen images during testing.

\subsection{Implementation Details}
Here we provide implementation details for both the proposed Exp-Eval and baselines.

\subsubsection{Exp-Eval Training}
To train the rating and explainer modules, we first generate explanations for the training data. For each dataset, we extract a small set of evaluation criteria (e.g., four criteria) from the examples and prompt GPT-OSS-120B to evaluate each training sample according to these criteria, producing a concise explanation for each criterion followed by a score on a scale from 0 to 1. The resulting explanations are collected and used as training data to fine-tune both the rating and explainer modules via SFT.

To train the rating module, the prompt consists of the input–output pair augmented with the corresponding explanations generated by GPT-OSS-120B, while the completion is the textual annotation of the ground-truth rating provided in the dataset. To train the explainer module via SFT, the prompt contains only the input–output pair, and the completion is the explanation generated by GPT-OSS-120B.

To improve generalization, we merge the training samples from all three datasets into a unified dataset and randomly permute the samples. We implement SFT using the TRL package. For each module, we train a LoRA adapter with hyperparameters $r=16$ and $\alpha=32$, and set the dropout rate to 0.05. Training is conducted for two epochs on approximately 70,000 samples with a batch size of 256 and a learning rate of $10^{-4}$. We use DeepSpeed \cite{deepspeed} to distribute training across four NVIDIA H200 GPUs.

To further improve the performance of the explainer module, we refine it via RLHF using the setting described in Subsection \ref{sec:gen-exp}. We fine-tune the base explainer module obtained via SFT to train the RLHF explainer. The RLHF explainer is implemented as a LoRA adapter with hyperparameters $r=16$ and $\alpha=32$, and set the dropout rate to 0.05. The activation function $\sigma(\cdot)$ is chosen as the sigmoid function. We employ GRPO from the TRL package to implement RLHF training. Training is conducted for two epochs with a batch size of 256 and a learning rate of $10^{-6}$.

To implement the ensemble model for Exp-Eval-ensemble, the learning rate $\eta$ is set to 0.01 across all tasks, and the exploration coefficient $\epsilon$ is set to 0.1. Ensemble predictions are obtained over 1,000 runs, and the averaged results are reported in Section \ref{sec:exp}. Notably, for the results in Table \ref{tab:2}, we use only a single run and perform the bootstrapping test on its predictions. For the image generation task, we did not observe significant performance differences between the selected run and the averaged performance across runs.

\subsubsection{Baselines}

To implement GenRM-CoT, we train a LoRA adapter via SFT to generate a chain of thought followed by the predicted score. GenRM formulates reward modeling as a next-token prediction task. We adopt brief explanation comments generated by GPT-OSS-120B as chain-of-thought supervision for GenRM-CoT. The LoRA hyperparameters are set to $r=16$ and $\alpha=32$, with a dropout rate of 0.05. The adapter is trained on the full training dataset for two epochs with a learning rate of $10^{-4}$ and a batch size of 256. The prompt consists of the input–output pair, and the completion is the corresponding chain-of-thought followed by the rating. The same settings are used for GenRM, except that it directly generates the score without producing a chain of thought. To evaluate the performance of all methods trained on datasets to generate CoTs or explanations, eight CoTs or explanations are generated for each sample, along with corresponding ratings. The results reported in the paper represent the average of these ratings to ensure consistency.

The prompt template used for the LLM-as-a-judge baselines is provided at the end of this appendix. Although this prompt does not include few-shot rated examples, we also evaluated the performance of the LLM-as-a-judge baselines with few-shot examples on the personalized recommendation task. The performance with few-shot examples was slightly worse than that without few-shot examples.

\subsubsection{Bootstrapping Details} \label{supp:boot}
We assess statistical significance using the bootstrap method with 10,000 resamples. Specifically, each bootstrap sample is generated by randomly sampling data points from the original dataset with replacement until a new dataset of the same size as the original is obtained. We then compute the correlation metric for each bootstrap sample. Finally, the 95\% confidence interval is estimated from the distribution of differences in correlation values across the 10,000 bootstrap samples. This procedure follows the standard bootstrap approach commonly used for estimating the uncertainty of evaluation metrics.

\subsubsection{Explanation Perturbation Protocol} \label{exp:pert}
Recall that explanations generated by Exp-Eval break down the evaluation task into a set of relevant criteria. To perturb the explanations, we randomly select a subset of criteria for each sample. Specifically, for samples from the Summarize from Feedback, HH-RLHF, and Goodreads datasets, we randomly select 1, 2, and 2 criteria, respectively. For each criterion in each dataset, we prepare a generic and nonspecific explanation that contains no sample-specific information. We then replace the explanation text associated with each selected criterion with its corresponding generic explanation. This procedure reduces the specificity of the explanations and may remove some helpful information. Consequently, the perturbed explanations are expected to be less useful while potentially retaining some useful information from the unperturbed portions.

\subsection{Supplementary Results} \label{supp:results}
Here, we provide supplementary experimental results and ablation studies for the proposed Exp-Eval.

\begin{table}[t]
\centering
\caption{Performance of Exp-Eval and baseline methods on RewardBench2.}
\label{tab:5}
\small
\setlength{\tabcolsep}{8pt}
\begin{tabular}{l||c|c|c|c|c|c|c}
\toprule
Method & Overall & Factuality & Precise IF & Math & Safety & Focus & Ties \\
\hline
GPT-4.1-2025-04-14 & 72.3  & 82.9 & 39.7 & 65.2 & 87.3 & 73.4 & 85.4 \\
Skywork-Reward-Gemma-2-27B-v0.2 & 75.3  & 76.7 & 37.5 & 67.2 & 96.9 & 91.7 & 81.8 \\
QRM-Gemma-2-27B & 76.7  & 78.5 & 37.2 & 69.6 & 95.8 & 95.4 & 83.2 \\
\hline
Exp-Eval & 76.6 & 76.6 & 50.6 & 71.3 & 86.2 & 92.1 & 82.7 \\
\bottomrule
\end{tabular}
\end{table}

\subsubsection{Generalization}
To assess the generalization ability of Exp-Eval-RLHF, we evaluate its performance on RewardBench tasks. Table \ref{tab:5} compares Exp-Eval with various baselines. Baseline results are taken from the RewardBench2 leaderboard. The backbone model used for Exp-Eval is Qwen3-14B \cite{qwen3technicalreport}. To better leverage the larger model capacity, we expand the training data for Exp-Eval by incorporating the HelpSteer3 dataset \cite{helpsteer3}. Although HelpSteer3 does not provide explicit numerical human ratings, it includes textual feedback that describes the helpfulness of each response. Specifically, the first line of each feedback entry describes the level of helpfulness and can be categorized into five distinct groups. We map these categories to Likert scores ranging from 1 to 5 and subsequently normalize the scores to the range [0, 1]. As shown, although Exp-Eval is trained as a specialized human-aligned evaluator for text summarization, AI assistant safety and helpfulness, and personalized recommendation, Exp-Eval achieves higher overall performance than GPT-4.1 indicating strong generalizability. Furthermore, Table \ref{tab:5} shows that Exp-Eval achieves higher overall performance than Skywork-Reward-Gemma-2-27B-v0.2 \cite{Liu2024skywork} and comparable to that of QRM-Gemma-2-27B \cite{Dorka2024}.

\subsubsection{Online Ensemble Learning}
Table \ref{tab:13} presents the results of ensembling Exp-Eval-RLHF with GPT-OSS-120B on the text summarization task. We use several model variants for Exp-Eval-RLHF. As shown, the ensemble method improves the Spearman correlation performance of Exp-Eval-RLHF across all model variants. In cases where Exp-Eval employs Qwen-3-4B and Gemma-2-9B, the ensemble approach achieves higher Spearman correlation than both Exp-Eval-RLHF and GPT-OSS-120B, indicating the effectiveness of online ensemble learning in leveraging the strengths of both models. Furthermore, when Exp-Eval uses Gemma-2-2B, there is a large performance gap between Exp-Eval-RLHF and GPT-OSS-120B. The ensemble model significantly improves the correlations of Exp-Eval while achieving performance close to that of GPT-OSS-120B. These findings are consistent with the theoretical analysis provided in Theorem \ref{th:1}.
\begin{table}[t]
\centering
\caption{Performance of Exp-Eval with and without ensembling across model variants on the text summarization task.}
\label{tab:13}
\small
\setlength{\tabcolsep}{46pt}
\begin{tabular}{l||cc}
\toprule
Method & $\rho$ & $\tau$ \\
\hline
GPT-OSS-120b & 0.4388 & 0.3737 \\
\hline
Exp-Eval-Gemma-2-2B & 0.1431 & 0.1287 \\
Exp-Eval-Gemma-2-2B-ensemble & 0.4375 & 0.3500 \\
\hline
Exp-Eval-Qwen-3-4B & 0.4358 & 0.3888 \\
Exp-Eval-Qwen-3-4B-ensemble & 0.4766 & 0.3831 \\
\hline
Exp-Eval-Gemma-2-9B & 0.4804 & 0.4336 \\
Exp-Eval-Gemma-2-9B-ensemble & 0.4857 & 0.4421 \\
\bottomrule
\end{tabular}
\end{table}

\begin{figure}[t]
    \centering
    \begin{minipage}{0.49\textwidth}
        \centering
        \includegraphics[width=\linewidth]{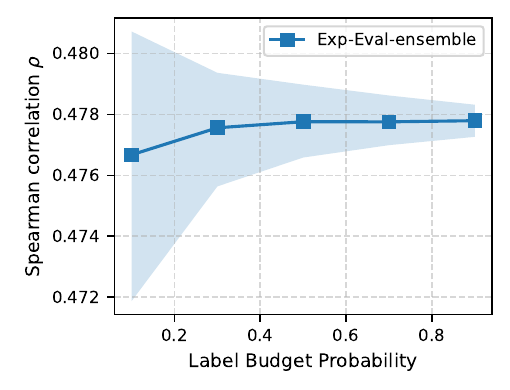}
        \small (a) Text Summarization
    \end{minipage}
    \hfill
    \begin{minipage}{0.49\textwidth}
        \centering
        \includegraphics[width=\linewidth]{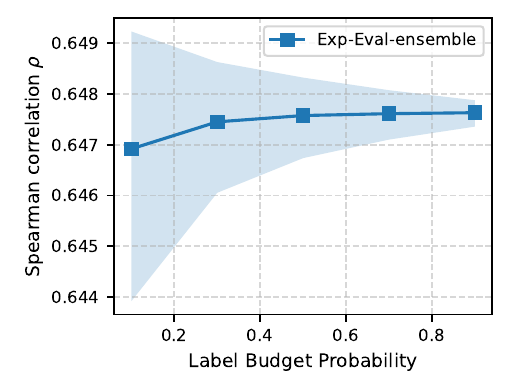}
        \small (b) AI Assistant Safety
    \end{minipage}
    \caption{Spearman correlation of Exp-Eval ensemble performance on text summarization and AI assistant safety tasks under varying label budgets.}
    \label{fig:ha_ab_study}
\end{figure}

Figure \ref{fig:ha_ab_study} examines the performance of Exp-Eval-ensemble under varying values of the label querying budget. The backbone model used for Exp-Eval is Qwen-3-4B.
The shaded regions denote 95\% confidence intervals computed from 1,000 ensemble runs.
As observed while increasing the budget leads to increase in Spearman correlation for both text summarization and AI assistant safety tasks, the variations in the correlations are relatively small. This demonstrates the robustness of the proposed ensemble approach. Furthermore, Figure \ref{fig:ha_ab_study} illustrates that as the label budget increases the variations in correlations across different runs drops dramatically. This observation matches with the intuition that higher randomization in observing labels can increase uncertainty in the performance of the model.

\section{Supplementary Related Works} \label{supp:related_works}
\textbf{Explanations and Reasoning}
Several recent methods have shown that generating critiques, reasoning traces, before producing a final judgment can improve both the quality and interpretability of evaluation and reward modeling \cite{Wei2022chain,madaan2023self,Akner2024,Yu2025,Trivedi2024,Wang2024self}. These approaches typically use a single large model to generate both critiques and scores, or train reward models that internally reason before outputting a score. Our work is complementary but distinct: we use explanations as a \emph{distillation interface} between a large teacher model and a small student evaluator. The explanations are not only an internal reasoning aid but a transferable representation that enables a lightweight model to acquire evaluation-relevant knowledge from a stronger model.

\textbf{Model distillation and teacher-student frameworks}
Distillation and teacher–student training are widely used to transfer knowledge from large models to smaller ones for efficiency \cite{hinton2015distilling, sanh2019distilbert}. In the context of LLMs, distillation has been applied to compress generative abilities, reasoning skills, or task performance \cite{hsieh2023distilling, magister2023teaching, ho2023large}. Exp-Eval can be viewed as a form of structured distillation, where the teacher provides criterion-level explanations rather than only soft labels or outputs. This structured signal enables the student not only to mimic the teacher’s scores, but to learn how to evaluate along dimensions that are meaningful for humans.

\textbf{Online learning and budgeted querying}
Our online ensembling component is related to work on online learning with expert advice \cite{littlestone1994weighted, freund1997decision, cesa2006prediction, agarwal2017corralling, Shen2019raker, foster2020instance} and partial or budgeted feedback \cite{cesa2006worst, dekel2012selective, Alon2015}, where a learner adaptively combines predictors while controlling query or labeling costs. We build on these ideas to adaptively mix ratings from a lightweight evaluator and a strong LLM-judge \cite{Zheng2023} under constraints on human and model queries. Unlike classical active learning \cite{settles2009active}, our setting does not focus on selecting training examples, but on controlling when and how expensive evaluators and labels are queried during deployment.

\textbf{Summary}
In contrast to prior work that either relies on fixed prompted judges or trains monolithic reward models, Exp-Eval combines (i) explanation-based supervision, (ii) lightweight task-adapted evaluators, and (iii) selective deferral to strong teachers into a single agentic framework. This combination allows us to achieve both high rating fidelity and robustness under distribution shift while maintaining practical efficiency.

\section{Limitations}
We discuss several limitations of the Exp-Eval framework. First, the maximum context length of evaluation tasks that Exp-Eval can handle is constrained by the context-length capacity of the student model. Applying Exp-Eval to tasks involving longer contexts therefore requires selecting student models with sufficient context-length capacity.

Second, Exp-Eval requires ground-truth ratings to train the rating module. The framework is primarily designed to improve automated evaluation for established, recurring tasks, such as AI Assistant Safety and Text Summarization, by leveraging both human-rated data and knowledge distilled from a teacher model. Although collecting human ratings incurs an upfront cost, the trained evaluation model can be reused for subsequent evaluations, potentially amortizing this cost over time. Nevertheless, the initial data collection requirement may limit the applicability of Exp-Eval to new tasks for which human-rated data are scarce or unavailable.

Third, the explanation usefulness and quality assessments in this paper rely on an LLM judge and lack human validation. Our perturbation analysis provides evidence that the judge is sensitive to reductions in explanation usefulness induced by replacing criterion-specific content with generic and nonspecific explanations. However, this analysis does not establish the extent to which the judge's assessments align with human judgments of explanation quality. Comprehensive human evaluation across multiple tasks and thousands of samples would require substantial annotation effort and is beyond the scope of this study. Therefore, our findings regarding explanation usefulness should be interpreted as assessments based on the specified LLM judge rather than as direct evidence of human-perceived usefulness. Nevertheless, the explainer module is primarily designed to support the rating module in producing human-aligned evaluations, and our explanation assessment provides complementary evidence regarding the quality of the generated explanations.

\section{Broader Impact} \label{broader}
Exp-Eval addresses the growing need for efficient system evaluation by accelerating the alignment of automated ratings with human judgment across diverse domains . While Exp-Eval automates evaluation tasks and reduces reliance on human annotators for efficiency, budgeted supervision improves performance and helps ensure reliability. The proposed approach can be used to efficiently monitor system performance with respect to user experience and satisfaction. At the same time, we recognize the dual-use nature of this research. Systems capable of approximating human judgments at scale could be misused if deployed without appropriate safeguards. For example, access to such technologies by malicious actors could enable the extraction of approximate population-level opinions on sensitive topics, which could then be leveraged for targeted persuasion or manipulation. To mitigate these risks, it is crucial to incorporate security, privacy, and access-control mechanisms into the deployment of such systems. We highlight the importance of carefully considering both the societal benefits and the potential unintended consequences of this work. We are optimistic that the broader impact of our contributions will lean toward positive and equitable outcomes.

\newpage
\newtcolorbox{promptllmbox}{
  colback=gray!5,
  title=Prompt Template for LLM-as-a-Judge Baselines,
  fonttitle=\bfseries,
  boxrule=0.5pt,
  arc=4pt,
  fontupper=\ttfamily\footnotesize,
}

\newtcolorbox{promptexamplebox}{
  colback=gray!5,
  title=Prompt Example for LLM-as-a-Judge Baselines,
  fonttitle=\bfseries,
  boxrule=0.5pt,
  arc=4pt,
  fontupper=\ttfamily\footnotesize,
}

\begin{promptllmbox}
You are an evaluator LLM. Your task is to assess the outcome or quality of a task based on the provided instructions, examples and inputs.

Your goal is to produce a final score that reflects human judgment.

---

Task Type: \{\textit{task type}\} \\
Task Instructions: \{\textit{task instructions}\} \\
Examples:\\
\{\textit{examples}\}

Input: \\
\{\textit{task input}\}

---

Task-Specific Instructions: \\
- \{\textit{scoring instructions}\}
\end{promptllmbox}

\begin{promptexamplebox}
You are an evaluator LLM. Your task is to assess the outcome or quality of a task based on the provided instructions, examples and inputs.
Your goal is to produce a final score that reflects human judgment.\\

---

Task Type: Assessment of User Ratings on Books \\
Task Instructions: You are given (1) a summary of a book’s content and characteristics, and (2) a summary of a user’s reading preferences, both derived from past reviews and ratings.

Your task is to estimate how much the user would like the book by analyzing how well the book aligns with the user’s preferences.\\

Input:\\
Summary of Reviews for the Book:\\
\{\textit{book summary}\}

User Preference Summary:\\
\{\textit{user preference summary}\}

---\\\\
Task-Specific Instructions:\\
- Estimate how much the user would like this book based on how well it matches the user's preferences.
Output a single score between 0 and 1.\\\\

Scoring guidelines:\\
- 1.0 = The user would almost certainly love the book (very strong match to preferences). \\
- 0 = The user would likely dislike the book (poor match). \\
- Intermediate values represent varying degrees of match. \\
Important:\\
- Use the full 0–1 range.
\end{promptexamplebox}

\newpage
\newtcolorbox{promptbox}{
  colback=gray!5,
  title=Prompt Template for Explanation Usefulness Assessment,
  fonttitle=\bfseries,
  boxrule=0.5pt,
  arc=4pt,
  fontupper=\ttfamily\footnotesize,
}

\begin{promptbox}
You are an expert evaluator responsible for determining whether a generated explanation satisfies minimum requirements for USEFULNESS.

Your task is to ACCEPT or REJECT the explanation.

This is a binary compliance judgment, not a quality rating.

========================
Acceptance Criteria
========================

Usefulness measures whether the explanation provides meaningful,
specific information that helps a human understand the relevant
evaluation or reasoning about the task.

ACCEPT only if:

- The explanation identifies relevant evidence, factors, or reasoning
  rather than merely making a generic statement.

- It provides enough substantive information for a human to understand
  what aspects of the task matter to the evaluation.

- The explanation connects the identified evidence or factors to the
  relevant criterion or task objective.

- The explanation provides information that would help a human
  independently evaluate or understand the reasoning.

- The explanation is sufficiently specific that it is distinguishable
  from a generic justification that could apply to many different
  tasks.

REJECT if:

- The explanation is substantially generic or superficial.

- It merely restates an assessment without explaining the relevant
  evidence or reasoning.

- It provides little or no information about what in the Task Input
  matters.

- Important reasoning needed to understand the explanation is absent.

- The explanation would provide essentially the same information if the
  Task Input were replaced with a different task.

- The explanation provides little practical value to a human trying to
  understand or evaluate the task.

Usefulness does NOT require:

- a particular conclusion;

- numerical scoring;

- exhaustive discussion of every piece of evidence;

- a particular writing style; or

- a long explanation.

A concise explanation can be highly useful if it identifies the key
evidence and explains its relevance.

========================
Important Instructions
========================

- Evaluate ONLY the generated explanation with respect to the Task
  Input and Task Instructions.

- Do not assume that there is a single correct overall evaluation
  unless the Task Input and Task Instructions explicitly establish one.

- Do not introduce external knowledge unless explicitly required by the
  task instructions.

- Do not reward or penalize the explanation based on its writing style,
  verbosity, or wording alone.

- Do not infer quality merely from whether the explanation "sounds
  convincing."

- Determine whether the explanation provides
  substantive, relevant information that helps a human understand the
  evaluation.

- When uncertain about whether a material requirement is satisfied,
  prefer REJECT.

- Give a brief, evidence-based justification for the decision.

========================
Output Format
========================

Return ONLY valid JSON:

{{
  "usefulness": {{

    "decision": "<accept|reject>",

    "justification": "<brief evidence-based explanation>"
  }}
}}

========================
Input
========================

Task Type:
\{\textit{task type}\}

Task Instructions:
\{\textit{task instructions}\}

Task Input:
\{\textit{task input}\}

Generated Explanation:
\{\textit{generated explanation}\}
\end{promptbox}

\end{document}